\documentclass{article} 
\usepackage{iclr2027_conference,times}
\usepackage[table]{xcolor}
\usepackage{wrapfig}

\usepackage{amsmath,amsfonts,bm}

\def\eqref#1{equation~\ref{#1}}

\def\1{\bm{1}}

\DeclareMathAlphabet{\mathsfit}{\encodingdefault}{\sfdefault}{m}{sl}
\SetMathAlphabet{\mathsfit}{bold}{\encodingdefault}{\sfdefault}{bx}{n}

\usepackage{hyperref}
\usepackage{url}
\usepackage{booktabs}
\usepackage{multirow}
\usepackage{enumitem}

\usepackage{graphicx}

\usepackage{algorithm}
\usepackage{algpseudocode}

\usepackage{enumitem}

\usepackage{pifont}  
\usepackage{xcolor}
\usepackage{natbib} 
\newcommand{\cmark}{\textcolor{green!60!black}{\ding{51}}}
\newcommand{\xmark}{\textcolor{red!80!black}{\ding{55}}}

\title{PowerBench: A Benchmark for Agentic Retrieval and Reasoning in Power Systems}

\author{\centering
Xijing Wang\(^{1}\) \thanks{Equal Contribution},
Yinsheng Yao\(^{2*}\) \thanks{The work was done when the authors studied at Tongji University.},
Jinru Ding\(^{3}\), Yidong Jiang $^{4\dagger}$,
Ziwen Xu$^{5\dagger}$, Yiwen Jiang$^{6}$, \and
\textbf{Jie Xu}\(^{3}\), 
\textbf{Dawei Cheng}\(^{1}\)\thanks{Corresponding author}
\\
\(^{1}\)Tongji University \quad \(^{2}\)Johns Hopkins University
\quad \(^{3}\)Shanghai Artificial Intelligence Laboratory  \\
\(^{4}\) Carnegie Mellon University\quad \(^{5}\)National University of Singapore \\
\(^{6}\) Big Data Center of State Grid Corporation of China
\\
\texttt{\{2353761, dcheng\}@tongji.edu.cn}
}

\iclrfinalcopy 
\begin{document}

\maketitle

\begin{abstract}
Large language model (LLM) agents offer new opportunities for automated analysis in industry.
However, rigorous evaluation of such agents-for example, within power system scenarios-remains hindered: real operational data are confidential, and existing public resources fail to fully capture the chained dependencies and heterogeneous evidence.
To address this gap, we propose PowerBench, comprising (1) a generation framework that derives interconnected heterogeneous operational data through a common dependency chain, and (2) a synthetic dataset generated by this framework.
The dataset covers 761 devices across 100 device types, with 13.35 million hourly telemetry records spanning two years and 24,939 operational documents.
Building on this dataset, we construct 300 questions across three task families that evaluate frontier LLMs' ability to complete analysis tasks that require autonomous evidence retrieval and reasoning across interconnected and heterogeneous data under restricted tool calls and time budgets.
Results demonstrate that the evaluated frontier LLMs remain challenged on these tasks: the best model reaches only 74.2\% joint accuracy. 
Our trace analysis further reveals that model performance varies across evidence discovery, content retrieval, tool use, reasoning over evidence, and answer submission. 
These findings provide detailed insights for evaluating LLM agents and guiding their reliable deployment in industry. 
The framework, dataset, and benchmark tasks are available at \url{https://github.com/open-compass/PowerBench}.
\end{abstract}

\section{Introduction}

Over the years, large language models (LLMs) have increasingly served as the foundation for autonomous agents that can plan, execute actions, and reason through complex tasks~\citep{plaat2025agentic,3780338.3783414,yehudai2026survey}.
This shift from simple conversational interfaces to agentic systems expands the capability boundaries and creates new opportunities for automated industrial analysis and task execution~\citep{jin2025gridmind}.
Retrieval and reasoning over heterogeneous data with logical links constitute a key challenge for deploying LLM agents in industrial settings~\citep{zheng2025planningarena,gao2023retrieval}.
Despite its importance, reliably evaluating these two abilities in real world industrial environments remains difficult.
On the one hand, confidentiality constraints restrict access to data collected from actual operations~\citep{zheng2022multi}; on the other hand, existing benchmarks fail to fully characterize the complexity of tasks in industry, accurately assess how retrieval and reasoning perform together, or provide detailed attribution of failures to specific stages~\citep{huang2025crmarena}.
Consequently, systematic benchmarks are urgently required to identify the inherent capability bottlenecks of current agents, support systematic failure analysis, and guide their reliable deployment in practice.

In real life power system scenarios, data inherently exhibit strong heterogeneity, encompassing multivariate time-series measurements and unstructured textual records such as operating logs and maintenance documents. 
Unlike conventional simple analysis tasks, power system workflows require multi-source evidence retrieval and multi-step logical reasoning along implicit dependency chains to support accurate diagnostic and analytical judgments~\citep{jia2025enhancing,tian2022event}.
For this reason, the power domain serves as a promising testbed for evaluating the retrieval and reasoning capabilities of LLM agents within large-scale, structurally complex data environments.

Despite this promise, existing public power datasets, including SGCC~\citep{zheng2017wide}, PowerGridQA~\citep{hannaan2026structured}, and Real-E~\citep{shao2025real}, are mostly limited to a single data modality and tailored to specific research tasks, such as electricity theft detection or power forecasting. 
Even recently proposed multimodal power datasets, such as MultimodalSyntheticPowerGrid~\citep{ni2026llm}, remain limited to pure data construction and lack executable agent tasks, and standardized evaluation protocols.

Beyond dataset limitations, current LLM evaluation benchmarks also exhibit notable shortcomings. 
Existing task-specific LLM agent benchmarks, including WorkArena~\citep{drouin2024workarena}, and Tau-Bench~\citep{26dddae50d5a4610bdaced3e9fdb1c89}, are constrained by simplified structural dependencies and low environment complexity, failing to simulate the intricate relational logic of real industrial tasks~\citep{huang2025crmarena}. 
Similarly, general multi-hop QA benchmarks such as HotpotQA~\citep{yang2018hotpotqa} and MuSiQue~\citep{trivedi2022musique} only support reasoning over pure textual sources. 
Collectively, these limitations prevent existing benchmarks from reliably evaluating industrial agent performance on retrieval and reasoning. 
We compare the characteristics of existing benchmarks and datasets with our proposed PowerBench in Table~\ref{tab:comparison}.

\begin{table}[t]
\centering
\setlength{\tabcolsep}{4pt}
\caption{
Comparison with representative datasets and benchmarks.
Data Connectivity: whether heterogeneous data sources have logical dependency or ordering relations, not just the same information in different forms.
Task Complexity: whether the resource defines agentic tasks requiring retrieval and reasoning.
Ground‑Truth Verifiability: whether agent outputs and actions can be programmatically validated against physical, logical, or operational constraints.
}
\label{tab:comparison}
\vspace{4pt} 
\begin{tabular}{lcccc}
\toprule
\textbf{Resource} 
& \shortstack{\textbf{Data}\\\textbf{Heterogeneity}} 
& \shortstack{\textbf{Data}\\\textbf{Connectivity}} 
& \shortstack{\textbf{Task}\\\textbf{Complexity}} 
& \shortstack{\textbf{Ground-Truth}\\\textbf{Verifiability}} 
\\
\midrule
SGCC                        
& \xmark & \xmark & \xmark & \xmark \\
PowerGridQA
& \xmark & \xmark & \xmark & \xmark \\
Real-E
& \xmark & \xmark & \xmark & \xmark \\
MultimodalSyntheticPowerGrid
& \cmark & \cmark & \xmark & \xmark \\
\midrule
WorkArena
& \cmark & \xmark & \cmark & \cmark \\
$\tau$-Bench
& \xmark & \xmark & \cmark & \cmark \\
HotpotQA
& \xmark & \xmark & \cmark & \cmark \\
MuSiQue
& \xmark & \xmark & \cmark & \cmark \\
\midrule
\textbf{PowerBench (Ours)}                         
& \cmark & \cmark & \cmark & \cmark \\
\bottomrule
\end{tabular}
\end{table}

To address these limitations, we construct and release PowerBench, a comprehensive benchmark tailored to evaluate the retrieval and reasoning capabilities of LLM agents on power system analytical tasks.
As illustrated in Figure~\ref{fig:powerbench}, PowerBench is built upon an LLM-assisted generation framework that produces structurally interconnected and heterogeneous data following unified logical dependency chains, alongside a large-scale synthesized power system dataset. 
Specifically, our benchmark tackles two key challenges:
(1) Object connectivity: it seeks to reproduce, as far as possible, the complex relationships between data objects.
(2) Task compositionality: it simulates multi-step power system analytical workflows, rather than oversimplified direct tasks such as web browsing and list filtering.
Our contributions are summarized as follows:
\begin{itemize}[leftmargin=*, topsep=2pt, itemsep=2pt, parsep=0pt, partopsep=0pt]
  \item We propose a data generation framework using LLMs to produce structurally interdependent, heterogeneous data through shared dependency chains, preserving consistency across sources.
  \item We release a large-scale multimodal power system dataset generated by this framework, alleviating the bottleneck in accessing real operational data.
  \item We build a comprehensive benchmark on the released dataset with three representative task families inspired by power system scenarios. It supports joint assessment of retrieval and reasoning capabilities and enables detailed attribution of failures to specific stages, thereby guiding the reliable deployment of LLM agents in practice.
\end{itemize}

\section{Related Work}
\textbf{Datasets for Power Systems.} 
Real operational data is difficult to access due to strict security constraints and regulatory restrictions in power systems~\citep{zheng2022multi,aravena2025open}.
Existing public power datasets alleviate this problem to some extent, but they are mostly limited to single data modalities and tailored for specific research tasks.
Specifically, SGCC~\citep{zheng2017wide} focuses on the detection of electricity theft and only releases real numerical records that exclude sensitive operational information.
PowerGridQA~\citep{hannaan2026structured} is constructed from publicly authoritative texts to evaluate the domain knowledge proficiency of large language models.
Real-E~\citep{shao2025real} aggregates publicly available ENTSO-E time-series data with its benchmark tasks focused on power forecasting.
Even the recently proposed MultimodalSyntheticPowerGrid~\citep{ni2026llm}, which generates heterogeneous data across images, text, and numerical modalities, remains limited to pure data construction.
Consequently, new benchmarks are needed to address the limitations of existing power system datasets, which rely on a single data modality, are designed for specific tasks, and lack standardized evaluation protocols.

\textbf{Benchmarks for LLM Agents.}
Prior studies have noted that most LLM agent benchmarks struggle to capture the structural interdependencies  of real world industrial environments, limiting their ability to measure practical retrieval and reasoning performance~\citep{huang2025crmarena}.
Traditional benchmarks, including HotpotQA~\citep{yang2018hotpotqa}, Tau-Bench~\citep{26dddae50d5a4610bdaced3e9fdb1c89} and MuSiQue~\citep{trivedi2022musique}, rely exclusively on pure textual corpora. 
Although they evaluate basic multi-document reasoning ability, they lack heterogeneous data sources and explicit cross-source dependency chains, which are essential for industrial analysis.
Recent agent benchmarks further advance task simulation but still suffer from fundamental structural defects. 
WorkArena~\citep{drouin2024workarena} provides multi-view webpage observations, yet all inputs originate from a single isomorphic webpage environment. 
These visual variations do not constitute heterogeneous industrial data with authentic logical or physical dependencies across different sources. 
Collectively, these benchmarks either restrict retrieval and reasoning to a single data modality or provide multimodal inputs without genuine heterogeneous cross-source dependencies, and they lack detailed attribution of failures to specific stages, thus failing to guide the reliable deployment of LLM agents in practice.

\textbf{Bottlenecks for Industrial Retrieval and Reasoning.}
For many industrial analysis tasks, reliable conclusions cannot be drawn from isolated data records~\citep{jia2025enhancing,tian2022event}.
Instead, agents must combine different categories of recorded information.
Nonetheless, current evaluation resources are inadequate to examine such capabilities, as summarized in existing benchmark research~\citep{zheng2025planningarena}.
First, most publicly released industrial datasets supply separate data entries, without explicitly annotating logical connections across related records~\citep{ni2026llm}.
Second, widely adopted LLM benchmarks assess retrieval and reasoning in isolation, which hinders diagnosis of failure points within end-to-end analytical pipelines~\citep{drouin2024workarena}.
This motivates PowerBench, which provides linked heterogeneous data and corresponding analytical tasks to systematically evaluate agentic retrieval and reasoning.

\section{Our PowerBench}
\label{sec:powerbench}
Our PowerBench, a benchmark built on a data generation framework, aims to address the limitations of current benchmarks for LLM agents in industrial analysis.
It integrates heterogeneous data, where they are interconnected through shared dependency chains, and defines structural analysis tasks calling for multi-step retrieval and reasoning, paired with verifiable ground truth for evaluation.
\begin{figure}[t]
\centering
\includegraphics[width=\linewidth]{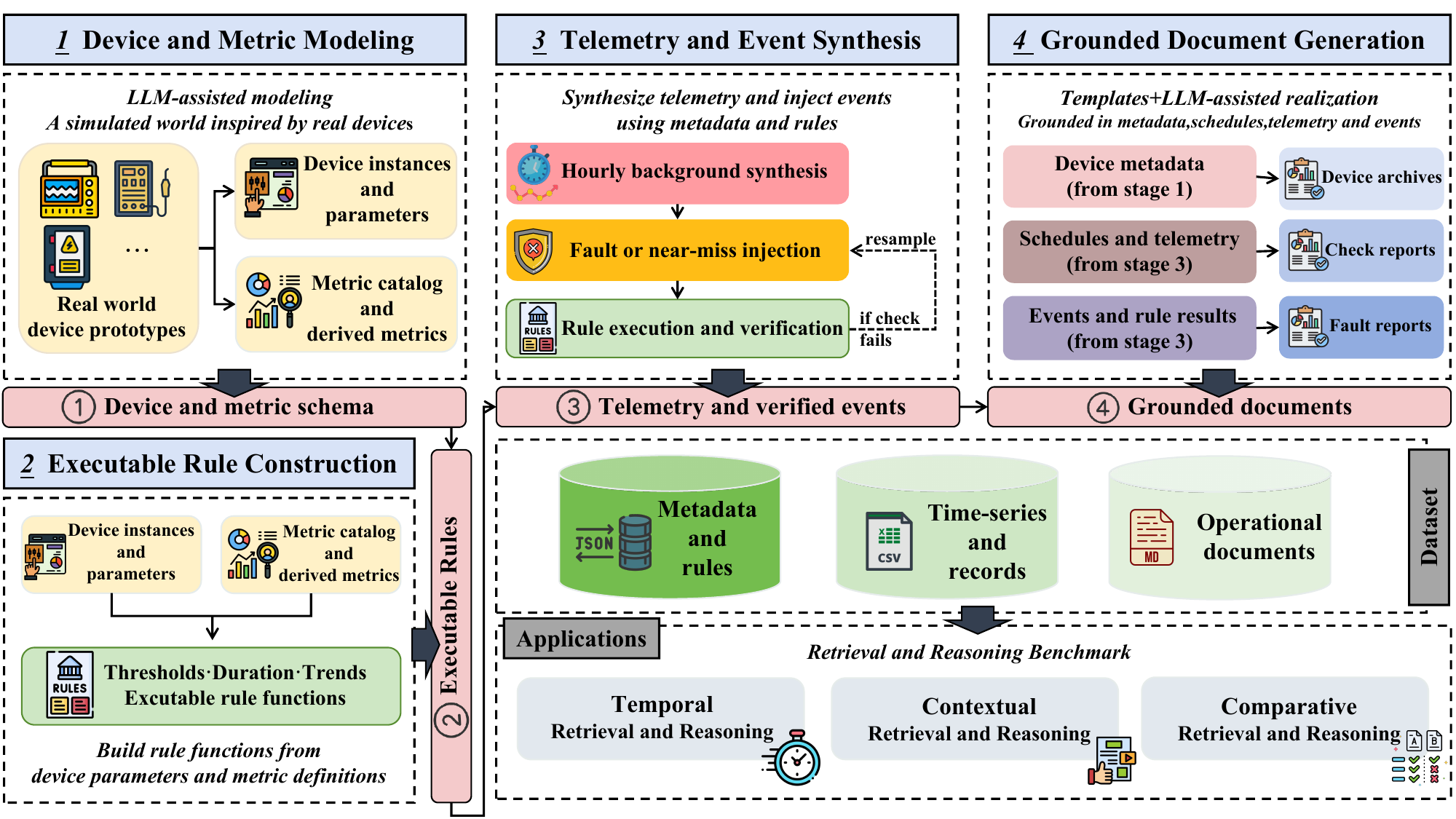}
\caption{Overview of PowerBench. All datasets come from the same structured state. Executable rules are used both to build the data and to verify the ground truth. The questions in the benchmark then ask models to retrieve and combine evidence from the world these rules produce, and to reason over that evidence.}
\label{fig:powerbench}
\end{figure}

\subsection{PowerBench Generation Framework}
\label{sec:dataset:construction}
As illustrated in Figure~\ref{fig:powerbench}, PowerBench organizes data around a device instance rather than around an isolated modality. For each device $d$, the framework exports:
\begin{equation}
\label{eq:device-bundle}
\mathcal{O}_d =
\left(
\mathcal{M}_d,\,
\mathcal{R}_d,\,
\mathcal{T}_d,\,
\mathcal{D}_d
\right).
\end{equation}
This contains four interrelated heterogeneous data that support our benchmark.
Specifically, $\mathcal{M}_d$ denotes the metadata of device $d$, including device configurations and metrics.
$\mathcal{R}_d$ represents a set of operational judgment rules relevant to the device.
$\mathcal{T}_d$ corresponds to hourly telemetry records.
$\mathcal{D}_d$ aggregates three categories of operational and maintenance reports.
They follow an explicit dependency structure:
\begin{equation}
\label{eq:four-source-dependency}
\mathcal{M}_d
\rightarrow
\mathcal{R}_d,
\qquad
(\mathcal{M}_d,\mathcal{R}_d)
\rightarrow
\mathcal{T}_d,
\qquad
(\mathcal{M}_d,\mathcal{R}_d,\mathcal{T}_d)
\rightarrow
\mathcal{D}_d.
\end{equation}

First,let $\mathcal{C}=\{c_1,\ldots,c_{|\mathcal{C}|}\}$ denote the manually curated taxonomy of equipment types. 
Each type $c\in\mathcal{C}$ has prototype $p_c$, described by its detailed information.
Given $p_c$, an LLM infers the dimensions $\mathcal{Z}_c$ along which instances of $c$ vary, and generates the corresponding instance set $\mathcal{I}_c$:
\begin{equation}
\mathcal{Z}_c = G_{\mathrm{dim}}(p_c),\qquad
\mathcal{I}_c = G_{\mathrm{inst}}(p_c,\mathcal{Z}_c).
\end{equation}
Both mappings are implemented by an LLM. 
The prototype fields and candidate dimensions are given in Appendix~\ref{app:device-expansion}.

For each generated device $d\in\mathcal{I}_c$, let $a_d$ denote its configuration. 
Let $\mathcal{K}_c$ and $\mathcal{H}_c$ denote the type-level atomic and derived metrics, respectively. 
The device metadata is
\begin{equation}
\label{eq:metadata-definition}
\mathcal{M}_d =
\left(a_d,\mathcal{K}_d,\mathcal{H}_d\right),
\qquad
\mathcal{K}_d\subseteq\mathcal{K}_c,
\quad
\mathcal{H}_d\subseteq\mathcal{H}_c,
\end{equation}
where $\mathcal{K}_d$ contains atomic metrics valid under $a_d$, and $\mathcal{H}_d$ the derived metrics computable from them. 
We audit the generated devices, configurations, and instance sets via web search; Appendix~\ref{app:metric-metadata} gives the procedure.

For each type $c$, the framework constructs a rule set $\mathcal{R}_c$. 
Each rule $r\in\mathcal{R}_c$ specifies the metrics it requires, its applicability conditions, and its threshold or temporal conditions. 
For a device $d$ of type $c$, we keep only rules that apply to its configuration $a_d$ and whose inputs are available:
\begin{equation}
\label{eq:device-rule-binding}
\mathcal{R}_d =
\left\{
r\in\mathcal{R}_c:
\operatorname{Applicable}(r,a_d)=1,
\operatorname{Inputs}(r)
\subseteq
\left(\mathcal{K}_d\cup\mathcal{H}_d\right)
\right\}.
\end{equation}
Here, $\operatorname{Applicable}(r,a_d)$ indicates whether rule $r$ applies to $a_d$, and $\operatorname{Inputs}(r)$ is the set of atomic or derived metrics needed by $r$. 
Thus $\mathcal{R}_d$ contains exactly the rules executable on device $d$. 
Appendix~\ref{app:rule-construction} gives the rule forms and construction details.

Given metadata $\mathcal{M}_d$ and executable rules $\mathcal{R}_d$, 
we generate nominal hourly background telemetry $X_d^{(0)}$ over time index $T$:
\begin{equation}
\label{eq:background-generation}
X_d^{(0)} =
B(T,\mathcal{K}_d,\mathcal{H}_d,a_d),
\qquad
\operatorname{Eval}(r,X_d^{(0)},a_d)=0,
\quad
\forall r\in\mathcal{R}_d.
\end{equation}
Here, $B$ generates background telemetry, and $\operatorname{Eval}$ returns 0 if rule $r$ is not triggered.
The second condition keeps $X_d^{(0)}$ nominal: no rule in $\mathcal{R}_d$ fires.
If violated, we resample the sequence.
Further details are given in Appendix~\ref{app:bg-eval}.

For a target event $e=(r,w)$, $r$ is the rule to trigger and $w$ its injection window. 
An injector $I_r$ modifies the background telemetry:
\begin{equation}
\label{eq:event-injection}
X_d^{(e)} =
I_r(X_d^{(0)},w,a_d),
\qquad
\operatorname{Eval}(r,X_d^{(e)},a_d)=1.
\end{equation}
Here $\operatorname{Eval}$=1 means the injected event triggers $r$.
We also generate near-miss events that leave the corresponding rule inactive. 
Let $E_d$ denote the retained fault and
near-miss events, and $L_d$ their schedules and verification records. 
Injecting all events in $E_d$ according to $L_d$ into $X_d^{(0)}$ gives the final telemetry $X_d$, so 
\begin{equation}
\label{eq:temporal-definition}
\mathcal{T}_d =
\left(X_d,E_d,L_d\right),
\end{equation}
Appendix~\ref{app:event-injection} gives the injection and near-miss details.

For device $d$, the document set is
\begin{equation}
\label{eq:document-definition}
\mathcal{D}_d =
\left(\mathcal{A}_d,\mathcal{C}_d,\mathcal{F}_d\right),
\end{equation}
where $\mathcal{A}_d$ is the device archive, $\mathcal{C}_d$ the inspection reports, and $\mathcal{F}_d$ the fault reports.  
These documents are grounded in the device metadata $\mathcal{M}_d$, rules $\mathcal{R}_d$, and temporal evidence $\mathcal{T}_d$:
\begin{equation}
\label{eq:document-grounding}
\mathcal{A}_d = G_{\mathrm{archive}}(\mathcal{M}_d,\mathcal{R}_d,\mathcal{T}_d),\quad
\mathcal{C}_d = G_{\mathrm{inspection}}(\mathcal{M}_d,\mathcal{T}_d),\quad
\mathcal{F}_d = G_{\mathrm{fault}}(\mathcal{M}_d,\mathcal{R}_d,\mathcal{T}_d).
\end{equation}
A fault report draws its event window, rule, thresholds, and telemetry observations from $\mathcal{R}_d$ and $\mathcal{T}_d$. 
This shared grounding makes document claims traceable to metadata, rule outcomes, and temporal evidence. 
Details are given in Appendix~\ref{app:document-generation}.

\subsection{PowerBench Dataset}
\label{sec:dataset:world}

\definecolor{tableheader}{RGB}{255,255,255} 
\begin{wraptable}{r}{0.5\textwidth} 
\centering
\setlength{\tabcolsep}{2pt}
\caption{Summary statistics of PowerBench.}
\begin{tabular}{ll}
\toprule
\rowcolor{tableheader} 
Statistic & Value \\
\midrule
Device Type & 100 \\
Device Instance & 761 \\
Metric & 1,275+203=1,478 \\
Rule & 2,044 \\
Telemetry & 13,350,984 \\
Metric Value & 207,878,856 \\
Report & 761+9,132+15,046=24,939 \\
Period & Jan.1, 2024-Dec.31, 2025 \\
\bottomrule
\end{tabular}
\label{tab:data}
\end{wraptable}

PowerBench is a large-scale heterogeneous and logically interconnected power system dataset generated by our proposed construction framework as shown in Table~\ref{tab:data}.
The dataset covers 100 distinct device types and 761 physical device instances.
For measurement metrics, it includes 1,275 native operational metrics together with 203 derived metrics, accompanied by 2,044 predefined fault and judgment rules to support downstream reasoning tasks.
The time-series telemetry subset contains 13,350,984 hourly records spanning from Jan. 1, 2024 to Dec. 31, 2025. 
Across all devices, this yields roughly 207.9 million individual metric values.
Every telemetry entry is tagged with consistent unique identifiers, enabling cross-source linkage between time-series measurements and textual operational documents.
The textual corpus consists of 24,939 operational documents in total: 761 device archives, 9,132 inspection reports, and 15,046 fault reports.
All data modalities are tied through a shared dependency chain, which supports agentic retrieval and multi-step reasoning.

\subsection{PowerBench Evaluation Task}
\label{sec:dataset:benchmark}
\begin{figure}[t]
\centering
\includegraphics[width=\linewidth]{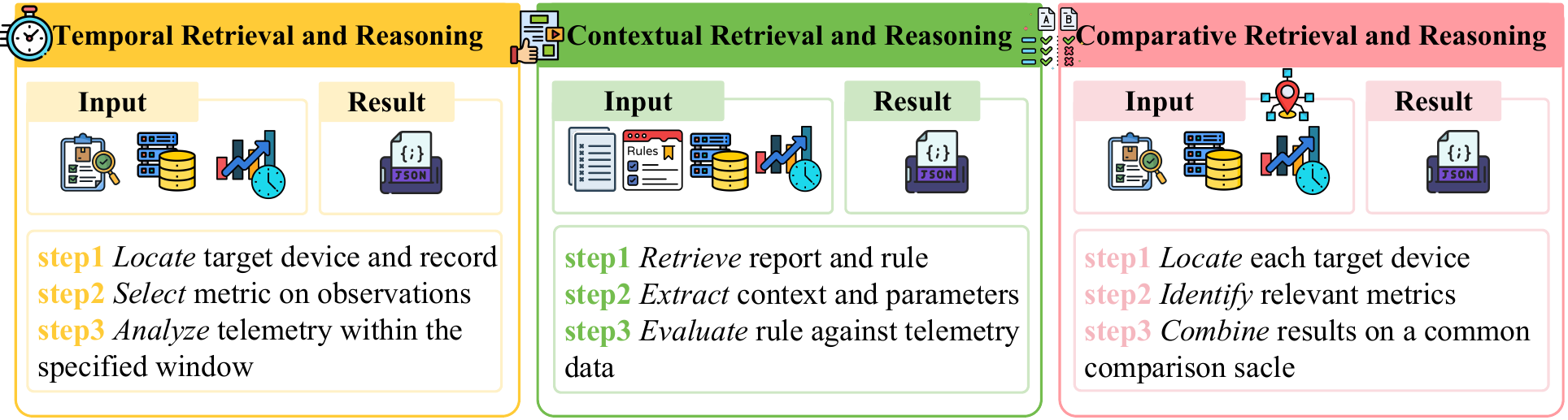}
\caption{The three task families in PowerBench: temporal, contextual, and comparative retrieval and reasoning. Each task requires the agent to retrieve heterogeneous evidence from the dataset and return a structured JSON answer and an evidence trace.}
\label{fig:task}
\end{figure}

Building on our dataset, we construct an agent benchmark to evaluate LLM agents’ ability to retrieve and reason over heterogeneous industrial evidence.
Each question gives only clues in natural language about a device or system phenomenon, with supporting evidence distributed across device metadata, judgment rules, hourly telemetry, and operational documents.
These sources supply device configurations and metric semantics, domain evaluation criteria, quantitative time‑series observations, and event context, respectively.
Solving a question thus requires retrieving and interpreting relevant records, computing via tool calls, and synthesizing information from multiple sources.
Under constrained tool calls and interaction budgets, PowerBench examines whether agents can leverage these interconnected heterogeneous data sources for industrial analysis workflows.
The task families are illustrated in Figure~\ref{fig:task}.
Appendix~\ref{app:taskQuestion} provides details of the data construction and task design.

\textbf{Temporal Retrieval and Reasoning Evaluation}
The Temporal Retrieval and Reasoning (TRR) task focuses on an LLM agent’s ability to diagnose a device after an event using both text and time-series evidence. 
It emphasizes turning descriptive observations into time-series analysis, rather than merely recalling static device knowledge. 
In practical scenarios, practitioners use inspection notes to identify metrics that need further investigation and then analyze subsequent telemetry to track how equipment condition changes. 
For the LLM agent, TRR requires retrieving the target device and its inspection report, selecting relevant metrics guided by documented observations and baseline values, and analyzing the corresponding telemetry readings. 
Inspection reports provide the context and starting timestamp; device metadata defines metric semantics and nominal baselines; telemetry records capture how equipment state evolves after the inspection. 
This task evaluates whether agents can turn free-text inspection findings into numerical analysis grounded in evidence.

\textbf{Contextual Retrieval and Reasoning Evaluation}
The Contextual Retrieval and Reasoning (CRR) task focuses on an LLM agent’s ability to reassess a device by applying rules under the operating context. 
It emphasizes using free-text event narratives to guide rule evaluation, rather than applying rules without context. 
In practice, system analysts revisit documented abnormal events and rerun judgment rules under revised adjustment policies, using the operating conditions recorded in fault reports. 
For the LLM agent, CRR requires retrieving fault reports and their matching device-level rules, extracting the operating context from fault narratives, choosing the applicable rule parameters, and evaluating the rules against available telemetry. Fault reports provide the context that determines parameter selection, while rules and telemetry together determine the evaluation result. 
This task evaluates whether agents can turn descriptive narratives into concrete conditions for rule evaluation.

\textbf{Comparative Retrieval and Reasoning Evaluation}
The Comparative Retrieval and Reasoning (CoRR) task focuses on an LLM agent’s ability to compare conditions across devices by combining evidence from multiple independent device instances. 
It emphasizes combining results from separate per-device analyses, rather than filtering a single set of similar records. 
In practice, maintenance practitioners gather inspection and telemetry information from multiple units to identify the devices that deviate most from normal operation. 
For the LLM agent, CoRR requires retrieving records for each target device, selecting relevant metrics, analyzing each device independently, and comparing the results on a common scale. 
Each device has its own documents, metric definitions, and temporal context. This task evaluates whether agents can combine separate per-device analyses while keeping each device tied to its own evidence.

\section{Experiments}
\label{sec:experiments}

\subsection{Evaluation Setup and Metrics}
\textbf{Evaluation Setup.} 
We evaluate GPT-5.6‑Sol~\citep{openai2026gpt56}, DeepSeek‑V4‑Pro~\citep{xu2026deepseek}, Qwen3.8‑Max~\citep{qwen2026max}, Kimi‑K3~\citep{team2026kimi}, GLM‑5.3~\citep{glm202653}, MiniMax‑M3~\citep{minimax2026m3}, Llama‑4‑Maverick~\citep{adcock2026llama}, Doubao‑Seed‑Evolving~\citep{bytedance2026doubao} and Gemini‑3.6‑Flash~\citep{googledeepmind2026gemini36} on the full benchmark of 300 questions.
All models receive the same system instructions and use the same tools through a shared client as shown in Table~\ref{tab:agent-tools}.

Each question is run under a fixed budget: 50 tool calls and 300 seconds for retrieval and reasoning. 
\definecolor{tableheader}{RGB}{233,150,122}
\begin{wraptable}[10]{r}[-25pt]{0.5\textwidth}
\vspace*{-8pt}
\centering
\setlength{\tabcolsep}{2pt}
\caption{Retrieval tools shared by all models.}
\begin{tabular}{ll}
\toprule
Name & Purpose \\
\midrule
\texttt{list\_corpus} & List corpus files \\
\texttt{search\_corpus} & Search paths, text, and headers \\
\texttt{read\_file} & Read Markdown or JSON files \\
\texttt{read\_timeseries} & Materialize CSV windows \\
\texttt{python} & Compute results \\
\bottomrule
\end{tabular}
\label{tab:agent-tools}
\end{wraptable}
Once either limit is reached, tools are disabled, and the model has one final chance to answer using only the evidence already collected. 
This final answer does not use additional tool calls but is included in the measured latency. 
Outputs are checked against the required JSON schema, with at most one automatic retry for formatting. Appendix~\ref{app:eval-details} details the full pipeline.

\textbf{Evaluation Metrics.} 
In assessing the correctness of the answer, our primary metric is Macro Joint Accuracy (Macro-JA), which averages Joint Accuracy (JA) equally across the three task families. 
JA treats a question as correct only if all of its answer fields are correct. 
To complement this strict metric, we also report Field Accuracy (FA), the average fraction of correct answer fields per question. 
We report JA and FA per task family, and Macro-JA and overall FA with 95\% bootstrap confidence intervals.
To locate retrieval failures before the final answer, we report Retrieved Evidence Recall (RER) and Available Evidence Recall (AER). 
RER is the fraction of required evidence sources returned by retrieval operations. 
AER is the fraction of required evidence content actually fetched into the agent's context. 
Both are computed per question and then averaged over the benchmark.

Beyond that, we use execution traces to report mean tool calls (TC), tool error rate (TER), and schema validity (VS). 
TC is the average number of tool calls per question, including failures. 
TER is the fraction of tool calls that return errors. 
VS is the fraction of final answers that pass the JSON schema required for each question, measuring parsability rather than correctness. 
We also derive two views from the same traces: Tool Reliability, $\mathrm{TR}=100-\mathrm{TER}$, and Tool Efficiency (TE), a relative score computed within each task family that assigns 100 to the model with the fewest mean tool calls. 
TE reflects cost only, not answer quality.
Exact definitions are provided in Appendix~\ref{app:metrics}.

\subsection{Overall Performance}

\definecolor{bestdark}{RGB}{255,210,220} 
\definecolor{secondlight}{RGB}{255,230,236}
\begin{table*}[t]
\centering
\setlength{\tabcolsep}{2.7pt}
\caption{Main results (in \%). Macro‑JA and overall FA include 95\% bootstrap confidence intervals. Best values among the evaluated models are marked with dark pink; second‑best values are marked with light pink.}
\label{tab:main_results}
\vspace{4pt} 
\begin{tabular}{lcccccccc}
\toprule
\multirow{2}{*}{\textbf{Model}} 
& \multicolumn{2}{c}{\textbf{TRR}} 
& \multicolumn{2}{c}{\textbf{CRR}} 
& \multicolumn{2}{c}{\textbf{CoRR}} 
& \multicolumn{2}{c}{\textbf{Overall}} \\
\cmidrule(lr){2-3} \cmidrule(lr){4-5} \cmidrule(lr){6-7} \cmidrule(lr){8-9}
& \textbf{JA} & \textbf{FA} 
& \textbf{JA} & \textbf{FA} 
& \textbf{JA} & \textbf{FA} 
& \textbf{Macro-JA [95\% CI]} & \textbf{FA [95\% CI]} \\
\midrule
GPT-5.6-Sol 
& \cellcolor{secondlight}85.0 & \cellcolor{secondlight}88.1 
& 88.9 & 93.5 
& \cellcolor{bestdark}\textbf{48.7} & \cellcolor{bestdark}\textbf{74.0} 
& \cellcolor{bestdark}\textbf{74.2 [69.5, 78.7]} & \cellcolor{bestdark}\textbf{82.7 [79.4, 85.8]} \\
GLM-5.3 
& 80.0 & \cellcolor{bestdark}91.9 
& \cellcolor{bestdark}\textbf{96.7} & \cellcolor{bestdark}\textbf{97.4} 
& \cellcolor{secondlight}13.3 & \cellcolor{secondlight}35.1 
& \cellcolor{secondlight}63.3 [59.2, 67.3] & \cellcolor{secondlight}65.2 [62.2, 68.1] \\
Doubao-Seed-Evolving 
& \cellcolor{bestdark}\textbf{86.7} & \cellcolor{bestdark}\textbf{94.4} 
& \cellcolor{secondlight}91.1 & \cellcolor{secondlight}95.6 
& 0.0 & 21.0 
& 59.3 [55.7, 62.6] & 58.1 [56.4, 59.6] \\
DeepSeek-V4-Pro 
& 50.0 & 68.3 
& 88.9 & 90.0 
& 0.7 & 13.8 
& 46.5 [41.7, 51.1] & 47.6 [44.5, 50.4] \\
Gemini-3.6-Flash 
& 55.0 & 67.2 
& 82.2 & 90.7 
& 2.0 & 10.0 
& 46.4 [41.3, 51.3] & 45.6 [42.6, 48.7] \\
Kimi-K3 
& 48.3 & 72.2 
& 85.6 & 94.4 
& 2.7 & 30.4 
& 45.5 [40.5, 50.5] & 58.0 [55.5, 60.5] \\
MiniMax-M3 
& 15.0 & 32.8 
& 62.2 & 73.9 
& 0.0 & 6.1 
& 25.7 [21.3, 30.4] & 31.8 [28.6, 34.8] \\
Qwen3.8-Max 
& 15.0 & 23.3 
& 40.0 & 63.0 
& 0.0 & 4.3 
& 18.3 [13.9, 23.0] & 25.7 [22.7, 28.7] \\
Llama-4-Maverick 
& 0.0 & 2.8 
& 6.7 & 36.5 
& 0.0 & 1.4 
& 2.2 [0.7, 4.1] & 12.2 [10.3, 14.2] \\
\bottomrule
\end{tabular}
\end{table*}
\vspace{-5pt} 

Table~\ref{tab:main_results} presents performance on the full 300-question benchmark, which covers 761 devices. 
Even the best model reaches only 74.2\% Macro-JA, and performance varies widely across models. 
DeepSeek-V4-Pro and Gemini-3.6-Flash have overlapping 95\% bootstrap confidence intervals, so their overall performance is not statistically different.

Across task families, performance varies widely. 
Many models perform reasonably well on the single-device TRR and CRR tasks. 
For example, Doubao-Seed-Evolving reaches 86.7\% JA on TRR, GLM-5.3 even reaches 96.7\% JA on CRR.
By contrast, CoRR, the multi-device comparison task, is much harder: the best model, GPT-5.6-Sol, reaches only 48.7\% JA, and most others are near zero. 
FA shows the same contrast: individual fields are often correct, but complete answers remain rare. 
For instance, on CoRR, GPT-5.6-Sol attains 74.0\% FA but only 48.7\% JA, and Kimi-K3 reaches 30.4\% FA but only 2.7\% JA. 
This gap indicates that strong single-device analysis does not easily transfer to tasks that require combining evidence across devices, which motivates PowerBench.

JA and FA, however, only measure final answer quality and do not show why a model fails. 
A model can fail for different reasons: it may never find the needed evidence; 
it may find a source but fail to retrieve usable content; 
tool calls may fail; 
reasoning may be wrong even with complete evidence; 
or the answer may not match the required schema. 
Most agent benchmarks only report final accuracy and cannot separate these failure modes. 
PowerBench keeps full execution traces, so we can diagnose these failures directly.

Moreover, PowerBench is not limited to the nine models evaluated here: it provides a reusable benchmark for diagnosing new LLM agents under the same evidence and budget constraints. 
We now turn to recurring failure sources: evidence access, tool errors, and incomplete answers.

\subsection{Evidence Access, Tool Errors, and Incomplete Answers}

\begin{figure*}[t]
  \centering
  \includegraphics[width=\textwidth]{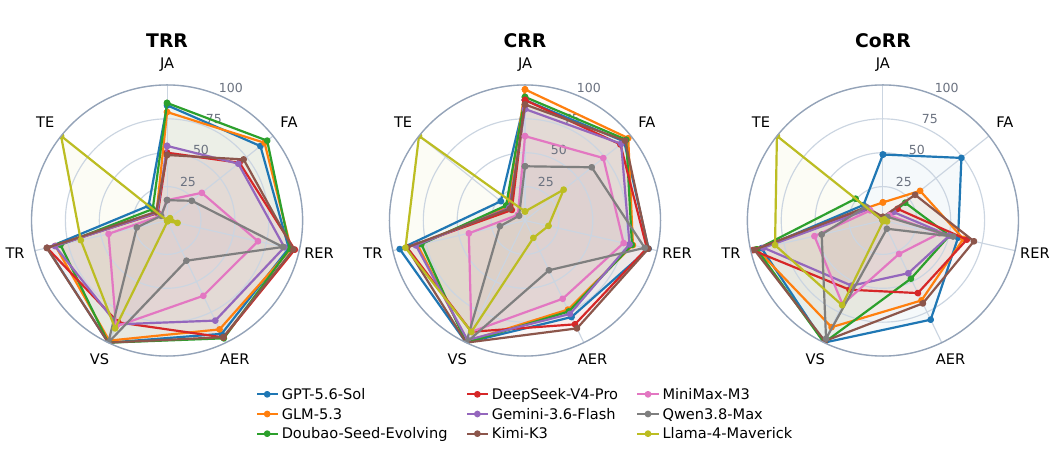}
  \caption{Multi-dimensional capability profiles across the TRR, CRR, and CoRR task families. Axes are scaled to \([0,100]\), higher is better.}
  \label{fig:radar_by_task}
\end{figure*}

PowerBench records full execution traces, so we can break each question into stages: 
whether the needed sources are found (RER), 
whether the required evidence is actually fetched (AER), 
whether tools succeed (TR $=100-\mathrm{TER}$), 
whether the final answer matches the JSON schema (VS), 
and whether the model reaches the correct answer (JA and FA). 
TE additionally summarizes relative cost of tool calls within each family.
Figure~\ref{fig:radar_by_task} shows these profiles by task family and Table~\ref{tab:radar_statistics_selected} lists the corresponding values.

\definecolor{bestdark}{RGB}{255,210,220} 
\definecolor{secondlight}{RGB}{255,230,236}
\begin{table*}[t]
\centering
\setlength{\tabcolsep}{0.3pt}
\caption{Statistics by task family for the radar plots.
JA and FA are reported in Table~\ref{tab:main_results}.
RER: retrieved evidence recall;
AER: available evidence recall;
VS: schema validity;
TR: tool reliability ($100-\mathrm{TER}$);
TE: task-wise relative tool efficiency, where the fewest mean tool calls score $100$ within each task family.
All values are percentages.
Best values per column are shaded dark pink and second-best values light pink.}
\label{tab:radar_statistics_selected}
\vspace{4pt}
\fontsize{8.5pt}{9pt}\selectfont
\begin{tabular}{@{}l ccccc ccccc ccccc@{}}
\toprule
\multirow{2}{*}{\textbf{Model}}
& \multicolumn{5}{c}{\textbf{TRR}}
& \multicolumn{5}{c}{\textbf{CRR}}
& \multicolumn{5}{c}{\textbf{CoRR}} \\
\cmidrule(lr){2-6} \cmidrule(lr){7-11} \cmidrule(lr){12-16}
& \textbf{RER} & \textbf{AER} & \textbf{VS} & \textbf{TR} & \textbf{TE}
& \textbf{RER} & \textbf{AER} & \textbf{VS} & \textbf{TR} & \textbf{TE}
& \textbf{RER} & \textbf{AER} & \textbf{VS} & \textbf{TR} & \textbf{TE} \\
\midrule
GPT-5.6-Sol
& 92.08 & 92.79 & \cellcolor{bestdark}100.00 & 90.80 & \cellcolor{secondlight}17.43
& \cellcolor{bestdark}93.52 & 79.07 & \cellcolor{bestdark}100.00 & \cellcolor{bestdark}94.77 & \cellcolor{secondlight}22.79
& 56.87 & \cellcolor{bestdark}81.21 & \cellcolor{bestdark}100.00 & 91.84 & 18.32 \\
GLM-5.3
& 93.83 & 89.15 & \cellcolor{secondlight}98.33 & 84.44 & 9.87
& 81.67 & 72.96 & \cellcolor{secondlight}98.89 & 81.28 & 16.37
& 60.90 & 65.47 & 87.33 & 95.42 & 14.96 \\
Doubao-Seed-Evolving
& 92.86 & \cellcolor{bestdark}96.46 & \cellcolor{bestdark}100.00 & 80.62 & 13.55
& 80.93 & 74.44 & \cellcolor{bestdark}100.00 & 78.38 & 17.79
& 50.25 & 47.60 & \cellcolor{bestdark}100.00 & 95.44 & \cellcolor{secondlight}25.78 \\
DeepSeek-V4-Pro
& \cellcolor{bestdark}96.61 & \cellcolor{secondlight}95.72 & 83.33 & \cellcolor{secondlight}90.77 & 8.75
& 92.22 & \cellcolor{secondlight}85.00 & 91.11 & 89.21 & 12.46
& \cellcolor{secondlight}63.48 & 59.43 & 56.67 & \cellcolor{bestdark}98.49 & 14.36 \\
Gemini-3.6-Flash
& 88.73 & 81.95 & 85.00 & 84.80 & 9.16
& 79.07 & 76.48 & 97.78 & 82.54 & 14.90
& 50.23 & 43.18 & 53.33 & 90.34 & 13.40 \\
Kimi-K3
& \cellcolor{secondlight}95.60 & 95.51 & \cellcolor{bestdark}100.00 & \cellcolor{bestdark}91.02 & 10.26
& \cellcolor{secondlight}93.33 & \cellcolor{bestdark}88.33 & \cellcolor{bestdark}100.00 & 89.47 & 14.10
& \cellcolor{bestdark}68.86 & \cellcolor{secondlight}67.86 & \cellcolor{secondlight}98.67 & \cellcolor{secondlight}97.27 & 15.80 \\
MiniMax-M3
& 68.77 & 61.65 & 86.67 & 44.28 & 5.76
& 74.63 & 64.07 & 91.11 & 42.50 & 6.69
& 42.74 & 27.40 & 68.67 & 51.80 & 11.88 \\
Qwen3.8-Max
& 87.66 & 32.92 & \cellcolor{bestdark}100.00 & 22.90 & 5.55
& 90.56 & 40.74 & \cellcolor{bestdark}100.00 & 18.98 & 5.60
& 53.08 & 6.75 & 96.67 & 46.34 & 16.23 \\
Llama-4-Maverick
& 7.91 & 0.73 & 88.33 & 65.28 & \cellcolor{bestdark}100.00
& 17.78 & 14.26 & 91.11 & \cellcolor{secondlight}90.43 & \cellcolor{bestdark}100.00
& 3.13 & 0.61 & 69.33 & 81.63 & \cellcolor{bestdark}100.00 \\
\bottomrule
\end{tabular}
\end{table*}

On TRR, evidence retrieval is relatively easy. 
Most models have RER and AER in the high 80s to mid 90s, meaning they can locate the target device from the inspection report and read the following telemetry window. 
Yet JA still varies widely. 
For example, DeepSeek-V4-Pro has RER 96.61\% and AER 95.72\% but only 50.0\% JA, while Doubao-Seed-Evolving reaches 86.7\% JA with AER 96.46\%. 
Finding evidence is therefore necessary but not enough: remaining errors arise in computation over windows, such as locating first occurrence times, and in composing the final answer. 


On CRR, performance is the strongest of the three families, suggesting that rerunning rules with fault report context is relatively easy. 
GLM-5.3 leads with 96.67\% JA and 97.41\% FA. 
Even here, finding a source often fails to translate into reading it. 
Qwen3.8-Max reaches 90.56\% RER but only 40.74\% AER, with TER 81.02\% ($\mathrm{TR}=18.98\%$): it finds many relevant sources but fetches little of the required evidence because tool calls frequently fail. 
MiniMax-M3 shows similar fragility at lower RER. The gap is milder but still present among stronger models: GPT-5.6-Sol leads RER at 93.52\% yet reaches only 79.07\% AER, while Kimi-K3 turns discovery into reading more effectively (AER 88.33\%). 

CoRR is the most revealing family.
RER falls for every model relative to TRR and CRR (typically into the $50$--$70\%$ range), so multi-device source discovery is harder than single-device lookup.
Beyond that shared drop, models fail for different reasons.
The strongest negative evidence for a pure retrieval story is Kimi-K3: it leads CoRR RER ($68.86\%$), ranks second in AER ($67.86\%$), and keeps VS~$98.67\%$ and TR~$97.27\%$, yet scores only $2.7\%$ JA.
DeepSeek-V4-Pro similarly has the highest tool reliability ($\mathrm{TR}=98.49\%$) but only $0.7\%$ JA; GLM-5.3 reaches AER~$65.47\%$ with TER~$4.58\%$ but only $13.3\%$ JA.
For models that already have abundant evidence and reliable tool use, the bottleneck is computation for each device, comparison across devices, and final aggregation once the evidence is already in the context.


Taken together, these profiles show different failure modes. 
Llama-4-Maverick makes very few tool calls and retrieves almost no evidence;
Qwen3.8-Max and MiniMax-M3 make many calls but have high tool error rates that block evidence retrieval; 
and models such as Kimi-K3 retrieve enough evidence and use tools reliably, while DeepSeek-V4-Pro also uses tools reliably, yet both still fail at cross-device reasoning.
The benchmark is open to the community for evaluating more models under realistic industrial analysis conditions.
We further study how performance and resource use change with the number of devices in Appendix~\ref{app:corr-scaling}, and how enlarging the visible device corpus affects CoRR on a fixed question subset in Appendix~\ref{app:retrieval-scope}.
Limitations and threats to validity are discussed in Appendix~\ref{app:limitations}.

\section{Conclusion}
In our work,we introduce PowerBench, an open benchmark for LLM agents, consisting of a data generation framework, a dataset, and three benchmark tasks. 
Experiments with several frontier LLMs show that they perform reasonably well on single-device tasks but struggle on analysis across multiple devices. 
Execution traces show that many failures occur after the needed evidence has been retrieved. 
PowerBench can locate failures at each stage of an agent pipeline and help guide improvements in planning, retrieval, and reasoning. 
By providing a controllable and reproducible setting with diagnostics at each stage, PowerBench can accelerate research on reliable agentic systems for industrial analysis, especially where real operational data are restricted.
Future work will expand data generation, study agent designs for reasoning over multiple complex instances, and extend task coverage to more realistic industrial analysis scenarios.


\subsection*{AI use statement}
In this work, we used generative AI tools for the generation of synthetic dataset, as our framework leverages LLMs to generate data.
We have not used generative AI tools for developing theoretical models or conceptual frameworks, formulating mathematical claims, providing critical ingredients for proving mathematical claims, proposing or refining hypotheses, designing or providing feedback on research methodology or experiments, supporting qualitative and thematic data analysis, or interpreting results; 
the remaining required disclosure tasks are not applicable to this work.
Additionally, we used AI for assisting in the polishing wording,implementing codes, and assisting with translation, cleaning and reformatting datasets.
We have reviewed all AI-assisted work. 
Specifically, we checked the LLM-generated synthetic data for quality, relevance, and correctness; and we verified and tested the AI-assisted code debugging for correctness and reproducibility.
We take responsibility for the final content of this work, including text, claims or artifacts produced with the aid of generative AI.

\subsection*{Ethics statement}

The dataset in this work is synthetic and is constructed from publicly available and searchable sources. 
It does not contain personal data, or confidential utility records, and no human-subject studies were conducted. 
We therefore do not foresee any ethics-review issue requiring escalation. 
At the same time, because the benchmark concerns critical infrastructure, we emphasize that its results should not be interpreted as deployment accuracy for any specific utility. Downstream users should exercise care when applying the benchmark or its findings in real operational settings.

\subsection*{Reproducibility statement}

To facilitate reproducibility, we provide an anonymous GitHub repository at
\url{https://github.com/open-compass/PowerBench} during the review phase. 
The repository contains the complete source code for this work. 
The full dataset is large: 761 devices, about 13.35M rows of hourly telemetry, 24,939 operational documents, a bundle of 300 questions, and traces left by model runs. 
So we will publish the complete dataset and the products of evaluation on our official public GitHub repository after review.
Detailed descriptions of the datasets, data processing steps, and evaluation protocols are provided in Appendix~\ref{app:framework}, Appendix~\ref{app:eval-details}, and Section~\ref{sec:experiments}.
For the LLM-generated data, the generation prompts, decoding parameters, filtering criteria, and quality-check procedures are described in Appendix~\ref{app:reproducibility} and implemented in the released code.
Random seeds and computational environments are also specified in the repository and Appendix~\ref{app:reproducibility}.
We hope these resources enable the community to reproduce and build upon our work.


\bibliographystyle{iclr2027_conference}
\bibliography{references}

\appendix
\section{The Details of the generation framework}
\label{app:framework}

\subsection{Device-Type Taxonomy, Prototype Schema, and Instance Expansion}
\label{app:device-expansion}

This appendix details Stage~1 of PowerBench:
$Z_c=G_{\mathrm{dim}}(p_c)$ and
$\mathcal{I}_c=G_{\mathrm{inst}}(p_c,Z_c)$.
In the released corpus, $|\mathcal{C}|=100$ equipment types
expand to $761$ device instances.

\subsubsection{Taxonomy $\mathcal{C}$ and prototype $p_c$}
\label{app:prototype}
Each type $c\in\mathcal{C}$ is a leaf of a manually curated four-level
hierarchy. The associated prototype is a five-field record
\[
p_c=(L_1,L_2,L_3,L_4,\mathrm{App}),
\]
where $L_1$--$L_4$ are hierarchical category, equipment class,
device-family name, and specification/model (the identity of $c$),
and $\mathrm{App}$ is the typical application context.
For $G_{\mathrm{dim}}$, the LLM receives the concatenated
device-type path:
$\mathrm{path}(p_c)=L_1\rightarrow L_2\rightarrow L_3\rightarrow L_4$
(the first four fields only).

\subsubsection{Dimension inference $G_{\mathrm{dim}}$}
\label{app:g-dim}
There is no closed global dimension vocabulary.
Given $\mathrm{path}(p_c)$, an LLM proposes a \emph{type-specific}
set $Z_c$ under a fixed two-class schema:
\begin{itemize}
  \item \textbf{Discriminative dimensions:} technical configuration axes
  that change subtype, monitoring focus, or diagnosis
  (e.g., voltage class, winding connection, cooling method,
  bushing type, structural form).
  \item \textbf{Instantiation dimensions:} ledger axes that place an
  instance in a concrete project/location
  (e.g., host project/station, system position, phase,
  primary/backup role, device ID).
\end{itemize}
Each dimension is a JSON object with name, description,
recommended value set, and a required or optional flag.
Prompt constraints require type-relevant axes only,
conservative engineering-realistic value sets,
omission of low-confidence dimensions,
and a ban on placeholder toponyms.
Both classes are then merged into
$Z_c=\{(\textit{name},\,\textit{description},\,\textit{value range})\}$
for Stage~2 (instantiation dimensions first; duplicate names dropped).
Illustrative candidates for a $\pm$800\,kV converter transformer include
discriminative axes
\{winding connection, cooling method, bushing type,
grid-side voltage, tap-changer mode\}
and instantiation axes
\{host project, system position, phase, unit ID, primary/backup\}.
Across the released corpus, frequently realized columns include
device ID, system position, primary/backup, phase, cooling method,
host substation, rated voltage, insulation medium, and interface or protocol fields---matching the examples in the main text while remaining type-dependent.

\subsubsection{Instance expansion $G_{\mathrm{inst}}$}
\label{app:g-inst}
Given the L4 type name and $Z_c$, a second LLM call produces an
engineering-plausible inventory $\mathcal{I}_c$ without enumerating the
full Cartesian product of value ranges.
The model must (i)~use exactly the input dimension names as columns
(plus a device-name field), (ii)~state combinatorial constraints before
listing devices, (iii)~compose instance names only from provided fields
and the type name, and (iv)~avoid placeholder site names.
\textbf{Examples from the released inventory.}
For $\pm$800\,kV high-end converter transformers (\texttt{0002}),
realized axes include system position $\{\mathrm{pole\,1},\,\mathrm{pole\,2}\}$,
phase $\{\mathrm{A,B,C}\}$, unit ID, primary/backup,
winding connection, cooling, bushing type, grid-side voltage, and
tap-changer mode ($|\mathcal{I}_c|=15$).
For OTN transport OSN~8800 (\texttt{0052}),
realized axes include system position, device ID, primary/backup,
line rate, service type $\{\mathrm{OTN},\mathrm{SDH},\mathrm{PTN}\}$,
slot count, and AC/DC power supply ($|\mathcal{I}_c|=16$).

\subsection{Atomic and Derived Metrics, Device Metadata, and Plausibility Audit}
\label{app:metric-metadata}

This appendix complements Stage~1 after instance expansion.
Given \(\mathcal{I}_c\), we construct type-level metric catalogs
and bind them to each device configuration, then audit the resulting
device models with web-search assistance.
In the released corpus, the $100$ types total
$1275$ atomic metrics, $203$ derived metrics, and $761$ devices.

\subsubsection{Type-level atomic catalog $\mathcal{K}_c$}
\label{app:atomic-catalog}
For each equipment type $c$, an LLM proposes a monitoring catalog from
the type path, instance dimensions, and inventory cues.
Metrics are partitioned into:
\begin{itemize}
  \item \textbf{Common metrics:} applicable to all instances of $c$.
  \item \textbf{Special metrics:} additional atoms gated by
  \emph{trigger conditions} on configuration fields
  (field name $+$ admissible value set), e.g., cooling-water
  temperature/pressure/flow only when
  $\textit{cooling method}=\text{forced-oil air-cooled}$.
\end{itemize}
The proposal is materialized into a typed catalog
$\mathcal{K}_c$ (name, snake\_case variable/DB field, dtype, unit,
discrete value set if any, and optionally a normal-range annotation).
Atomic names exclude temporal aggregates (rolling means, $1$h/$24$h
windows, etc.); those belong to later temporal evaluation, not to
$\mathcal{K}_c$.

\subsubsection{Type-level derived set $\mathcal{H}_c$}
\label{app:derived-metrics}
Derived metrics are \emph{single-timestep, multi-atom} composites
(ratios, differences, physically motivated indices), proposed from
$\mathcal{K}_c$ and validated by a second LLM pass (accept/reject).
Hard constraints include: at least two atomic inputs; all atoms must
exist in $\mathcal{K}_c$; no time-window statistics
(trend/variance/range); no arithmetic over discrete status fields;
no arbitrary scoring/MinMax indices without an engineering basis.
Accepted entries form $\mathcal{H}_c$ (name, key, formula, units,
prerequisites, optional references such as IEC/DL oil-DGA ratio practice).

\subsubsection{Configuration-conditioned binding
$\mathcal{M}_d=(a_d,\mathcal{K}_d,\mathcal{H}_d)$}
\label{app:metric-binding}
For $d\in\mathcal{I}_c$, let $a_d$ be the instance attribute vector
(inventory columns: project/station, system position, cooling method,
protocol, \ldots). Binding is deterministic:
\begin{align}
\mathcal{K}_d
&=
\mathcal{K}_c^{\mathrm{common}}
\cup
\bigl\{
m\in\mathcal{K}_c^{\mathrm{special}}
:
\operatorname{Triggers}(m)\subseteq a_d
\bigr\},
\\
\mathcal{H}_d
&=
\bigl\{
h\in\mathcal{H}_c
:
\operatorname{Atoms}(h)\subseteq\mathcal{K}_d
\bigr\}.
\end{align}
That is, special atoms fire only when \emph{all} of their trigger
clauses match $a_d$; a derived metric is retained only if every
required atom is already in $\mathcal{K}_d$ (computability).
Device metadata is then
$\mathcal{M}_d=(a_d,\mathcal{K}_d,\mathcal{H}_d)$,
written onto per-type device sheets used by later rule binding and
telemetry synthesis (non-$\,\mathcal{K}_d$ channels are masked at generation time).

\subsubsection{Web-search plausibility audit}
\label{app:device-audit}
After expansion, we run an LLM auditor with web search over each of the $100$ taxonomy rows and its instance inventory.
The auditor looks up public material---national and industry standards (GB/T, DL/T, IEC), utility design notes, UHV project documents, and vendor literature---and checks whether the stated product or engineering designation, the four-level hierarchy, and the main application look consistent with that material; 
for the expanded inventory it also checks identity fields, value ranges, project/station naming, and rough instance counts.
Metric catalogs and executable rules are not part of this pass.
The point of the audit is modest.
PowerBench is inspired by real equipment and operating practice,not a release of confidential utility records, 
so we use searchable sources only to support that the modeled types and instances are engineering-plausible under public information.
A clean audit does \emph{not} mean the synthetic world reproduces on-site physics, protection settings, or private station ledgers, which utilities do not publish in full.

\subsection{Executable Operational Rules: Forms, Construction, and Binding}
\label{app:rule-construction}
This appendix details Stage~2 of PowerBench: construction of the
type-level rule set $\mathcal{R}_c$ and device binding of
$\mathcal{R}_d$ in Eq.~\eqref{eq:device-rule-binding}.
In the released corpus, the $100$ types yield $2044$ rules in total.

\subsubsection{Construction of $\mathcal{R}_c$}
\label{app:rule-gen}
Rules are generated by a two-stage LLM procedure that follows the Stage 1 metric catalog.

\textbf{Stage A (coverage objects).}
From the type path, dimensions, and inventory, the model enumerates
(i)~common metrics, (ii)~special metrics with configuration triggers, and
(iii)~special operational concerns.
This stage defines \emph{what} must be monitored; it does not yet emit
executable thresholds or formulas.

\textbf{Stage B (executable rules).}
For each Stage A object, the model emits one or more judgment rules.
Each $r\in\mathcal{R}_c$ is a structured record with the fields in
Table~\ref{tab:rule-schema}.
Optional post-processing can add further rules whose inputs are
derived metrics in $\mathcal{H}_c$, using the same schema.

\begin{table}[t]
\centering
\small
\caption{Schema of a type-level rule $r\in\mathcal{R}_c$.}
\label{tab:rule-schema}
\begin{tabular}{lp{0.58\linewidth}}
\toprule
Field & Role \\
\midrule
Rule ID & Identifier within type $c$ (e.g., K1, K12) \\
Target object & Linked Stage A metric or concern \\
Object class & Common metric / special metric / special concern \\
Applicability & $\operatorname{Applicable}(r,a_d)$:
  empty $=$ all instances of $c$; else AND of
  (config field $\in$ value set) clauses \\
Inputs & $\operatorname{Inputs}(r)$: atomic and/or derived metric names \\
Point condition & Whether a single sample may fire; consecutive-point count \\
Temporal condition & Optional window length and statistics
  (mean, max, slope, count, duration, \ldots) \\
Expression & Boolean DSL over metrics, windows, and named thresholds \\
Thresholds & Named numeric parameters referenced by the expression \\
\bottomrule
\end{tabular}
\end{table}

Generation constraints require concrete numeric thresholds(no ``TBD'' placeholders), coverage of all Stage A common and special
metrics, and consistency between consecutive-point / window
annotations and the operators appearing in the expression.

\subsubsection{Rule expression language}
\label{app:rule-dsl}
Expressions are evaluated on hourly telemetry.
The surface language supports Boolean connectives
($\mathrm{AND}$/$\mathrm{OR}$/$\mathrm{NOT}$), comparisons
($>,\ge,<,\le,=,\neq$, and membership $\mathrm{in}\{\ldots\}$),
arithmetic on metric channels, and windowed operators.
Named thresholds (e.g., $T_{\mathrm{alarm}}$, $R_{\mathrm{discharge}}$)
are substituted from the rule's parameter list.
At runtime, expressions are either compiled into typed Python checkers
or interpreted by a restricted AST evaluator that resolves Chinese
metric names via the catalog and computes any inline derived quantities.

\subsubsection{Device binding $\mathcal{R}_d$}
\label{app:rule-binding}
Given device metadata $\mathcal{M}_d=(a_d,\mathcal{K}_d,\mathcal{H}_d)$,
binding is deterministic:
\begin{enumerate}
  \item $\operatorname{Applicable}(r,a_d)=1$ iff every applicability
  clause matches $a_d$ (exact string membership); an empty clause list
  means globally applicable on type $c$.
  \item $\operatorname{Inputs}(r)\subseteq(\mathcal{K}_d\cup\mathcal{H}_d)$
  iff every listed metric is available after configuration-conditioned
  metric binding.
\end{enumerate}
Only rules satisfying both conditions enter $\mathcal{R}_d$ and are
written onto the device sheet for later telemetry synthesis.
A second eligibility pass at generation time re-checks applicability
under the device's field overrides and retains only rules for which an
executable injector and required data channels exist.
Thus $\mathcal{R}_d$ is exactly the set of rules that can be evaluated
on device $d$.
Corpus-level breakdown: of $2044$ type-level rules,
$866$ target common metrics, $679$ special metrics, and
$499$ special concerns; $1187$ have empty applicability
(global on the type) and $857$ are configuration-gated;
$748$ declare non-empty long-window temporal requirements.


\subsection{Nominal Background Telemetry Generation}
\label{app:bg-eval}

This subsection fills in Stage~3 background synthesis for
Eq.~\eqref{eq:background-generation}: how
$X_d^{(0)}=B(T,\mathcal{K}_d,\mathcal{H}_d,a_d)$ is built so that no
rule in $\mathcal{R}_d$ fires on the background, and how we resample when
that check fails.
Each of the $761$ devices has $|T|=17{,}544$ hourly samples
(2024--2025).

The time grid is
\[
T
=
\{t_0,t_0+\Delta,\ldots,t_1\},
\qquad
\Delta=1\,\mathrm{h},
\]
with default $t_0=\texttt{2024-01-01\,00:00}$ and
$t_1=\texttt{2025-12-31\,23:00}$.
Configuration $a_d$ is passed to both $B$ and $\operatorname{Eval}$.
Only fields in $\mathcal{K}_d$ are written as atomic channels; other
columns are masked.
Derived metrics in $\mathcal{H}_d$ are recomputed from atoms at
evaluation time (and when exporting the series), not filled by separate
noise.

\subsubsection{Background function $B$}
\label{app:bg-B}

Generation uses a shared fallback fill, then a type-specific generator
guided by the metric catalog and Stage~2 thresholds.

\textbf{Numeric plan.}
For each continuous atom $m\in\mathcal{K}_d$, we set a quiet band
$(b,\sigma,\ell,u)$ by (i)~taking the catalog normal range when valid,
(ii)~cutting the band to the safe side of Stage~2 alarm comparisons after
substituting named thresholds, (iii)~shrinking $[\ell,u]$ by about
$10\%$ per side, and (iv)~capping $\sigma\le(u-\ell)/6$ with $b$ inside
the band.

\textbf{Stochastic fill.}
Most types use mean-reverting AR(1) paths
\[
x_0=b+\varepsilon_0,\qquad
x_t=b+\phi(x_{t-1}-b)+\eta_t,\qquad
\phi=0.985,\quad
\eta_t\sim\mathcal{N}\!\bigl(0,\sigma\sqrt{1-\phi^2}\bigr),
\]
clipped to $[\ell,u]$.
Other channels use i.i.d.\ Gaussians, slow counters, near-flat walks
(e.g., SF$_6$ pressure), or light diurnal sinusoids (e.g., coolant
temperature).
Discrete/status atoms stay at a non-alarm token outside Stage~2 alarm
lists.

\textbf{Configuration-aware pins.}
When $a_d$ encodes ratings or subtypes, $B$ may overwrite generic draws
(e.g., load current $\approx 0.45\,I_{\mathrm{rated}}$; coupled open/close
times so a derived deviation ratio stays below its rule threshold).
This keeps short $\mathrm{consec}_{Nh}$, slope, and ratio rules quiet
without depending only on rejection sampling.

\subsubsection{Background cleanliness check and resampling}
\label{app:bg-clean}

After filling $X_d^{(0)}$, we form an eligible rule set
$\mathcal{R}_d^{\mathrm{elig}}\subseteq\mathcal{R}_d$ by intersecting
applicability, registry membership, injectability, and data availability.

We keep the background nominal in the sense of
Eq.~\eqref{eq:background-generation} in two practical ways:
\begin{enumerate}
  \item \textbf{Quiet bands and pins}
  (Appendix~\ref{app:bg-B}) keep continuous channels away from alarm
  thresholds and hold discrete tokens off alarm values, so the common
  short-window and threshold rules stay inactive over the written series.
  \item \textbf{Probe and resample.}
  We evaluate eligible rules at $n{=}6$ times spaced through the interior
  of $T$ (uniformly in $[0.15,0.95]$ of the index).
  If any rule fires at any probe, we regenerate with a deterministic
  reseed for up to three attempts---matching the main-text ``if violated,
  we resample.''
\end{enumerate}
The six probes are an acceptance check for that resample rule, not a
proof that every rolling-window statistic is quiet at every hour.
Released device reports that carry the standard check fields mark
background cleanliness as passed under this screen.

\subsubsection{Illustrative vignettes}
\label{app:bg-examples}

\textbf{MV switchgear.}
AR(1) fills plus rating-aware pins
($I_{\mathrm{load}}$, PD level, humidity, interlock/status) and open/close
time coupling keep consecutive over-current and operating-time ratio rules
quiet.

\textbf{MMC converter valve.}
Current/voltage stay near rated set-points; coolant has a mild diurnal
swing and body temperature tracks coolant with a bounded rise, so
relative temperature-rise rules stay quiet until later injection.
Quiet bands plus probe-and-resample therefore implement the nominal
background required by Eq.~\eqref{eq:background-generation}.

\subsection{Event Injection, Near-Misses, and Final Telemetry}
\label{app:event-injection}

This appendix details Stage~3 event synthesis after background generation:
how a target event $e=(r,w)$ is realised by an injector $I_r$, how
near-miss counterparts keep $\operatorname{Eval}=0$, and how retained
events assemble into
$\mathcal{T}_d=(X_d,E_d,L_d)$
(Eqs.~\eqref{eq:event-injection}--\eqref{eq:temporal-definition}).

\subsubsection{Injectors $I_r$ and verification}
\label{app:injector}

For each executable rule $r\in\mathcal{R}_d$ that admits an injector,
the per-type kit provides a paired function $I_r$ that edits a copy of
$X_d^{(0)}$ inside a lookback window ending at decision time
$t_e=\mathrm{end}(w)$.
Typical edit primitives include:
painting $N$ consecutive hours above/below a threshold
(for $\mathrm{consec}_{Nh}$ rules);
setting discrete status tokens;
linear ramps over multi-day windows (for slope rules);
and trailing ``bad'' hours inside a longer count window.
Thresholds are taken from the rule's named parameters
(with safe numeric fallbacks only when needed).
Derived channels are stripped before evaluation and recomputed from
atoms, so ratio/product rules see consistent inputs.

Acceptance follows Eq.~\eqref{eq:event-injection}: we require
\[
\operatorname{Eval}(r,I_r(X_d^{(0)},w,a_d),a_d)=1
\]
at $t_e$ (and not marked skipped).
Failed attempts are logged and left out of $E_d$.
As in the main text, this only checks that the intended rule fires; it
does not claim that no other rule in $\mathcal{R}_d$ also becomes active.

\subsubsection{Scheduling windows $w$ and records $L_d$}
\label{app:schedule}

On the same hourly grid $T$ as the background
($|T|=17{,}544$ by default), the orchestrator builds:

\begin{itemize}
  \item \textbf{Inspection days:} $12$ bi-monthly anchors (default),
  used mainly for near-miss placement;
  \item \textbf{Episode days:} $8$ sampled calendar days hosting
  fault clusters;
  \item \textbf{Per episode:} $2$--$4$ alerts, drawing from the
  eligible injector pool under a mutex that forbids overlapping
  occupied intervals (lookback length parsed from the rule's
  consecutive/window operators; long-horizon rules occupy at least
  the episode span).
\end{itemize}

Each retained event stores
$(r,\,w_{\mathrm{start}},\,w_{\mathrm{end}},\,\textit{type},\,
\textit{changed columns},\,\ldots)$
in a per-device JSON schedule---this is $L_d$, together with
inspection/episode calendars and post-hoc verification flags.
RNG streams are salted by model name and device index so identical
CLI seeds do not clone calendars across devices.

\subsubsection{Near-miss construction}
\label{app:near-miss}

A near-miss for rule $r$ uses the \emph{same} injector $I_r$, then
backs off toward the background so that the rule stays inactive.
Let $X^{\mathrm{base}}$ be the pre-injection series and
$X^{\mathrm{inj}}=I_r(X^{\mathrm{base}},w,a_d)$ a verified trigger.
For continuous channels,
\[
X^{(\alpha)}
=
(1-\alpha)\,X^{\mathrm{base}}
+
\alpha\,X^{\mathrm{inj}},
\qquad
\alpha\in(0,1],
\]
while discrete/status overwrites are applied only for larger $\alpha$
(threshold $0.5$ in the implementation).
The orchestrator searches a discrete $\alpha$-grid
(negative powers of two and $i/32$ fractions), optionally refines by
binary search, and retains
\[
\alpha^\star
=
\max\{\alpha:\operatorname{Eval}(r,X^{(\alpha)},a_d)=0\}.
\]
Thus near-misses lie close to the fault manifold without firing $r$.
In the released corpus, successful near-miss $\alpha$ values have
median $\approx 0.66$ (p10--p90 $\approx 0.50$--$0.93$).
Candidates that would re-trigger prior near-misses, violate mutexes,
or fail the clean-base precondition are skipped.

\subsubsection{Composing $E_d$ into final $X_d$}
\label{app:compose}

Events are merged sequentially onto one working frame:

\begin{enumerate}
  \item Start from clean $X_d^{(0)}$.
  \item Inject episode faults; accept only $\operatorname{Eval}=1$
  for the target rule (Eq.~\eqref{eq:event-injection}).
  \item Top up faults toward the soft density target on free slots
  (same mutex).
  \item Inject near-misses with $\alpha$-backoff
  ($\operatorname{Eval}=0$ for the \emph{corresponding} rule, as in
  the main text).
  \item Repair up to a few rounds: re-evaluate retained faults;
  re-inject any destroyed by later overwrites; resolve same-metric
  long-window conflicts by keeping the longer/earlier lookback.
  \item Rematerialise derived metrics and export.
\end{enumerate}

The retained fault and near-miss sets form $E_d$;
schedules plus per-event verification form $L_d$;
the final hourly series is $X_d$.
Hence $\mathcal{T}_d=(X_d,E_d,L_d)$.
Labels live in JSON (not as CSV columns); downstream document
generation and QA join events by device, rule ID, and
$w_{\mathrm{end}}$.
Repair restores intended target-rule semantics under composition;
it is not offered as a proof that unrelated rules stay silent.

\subsubsection{Corpus scale and checks}
\label{app:event-stats}

Across $761$ devices, retained events average about $34$ faults and
$10$ near-misses per device (about $2.6\times 10^{4}$ faults and
$7.4\times 10^{3}$ near-misses in total).
Each device report records three checks: background cleanliness under
the probe screen; every retained fault still triggers its target rule at
$t_e$; every retained near-miss leaves its corresponding rule inactive.
About $96.5\%$ of devices pass all three under the standard report
fields; the rest are incomplete exports rather than accepted
counterexamples to
Eqs.~\eqref{eq:background-generation}--\eqref{eq:event-injection}.
Fault/near-miss density targets are soft.

\textbf{Benchmark labels.}
These corpus checks are not the gold oracle for questions.
For each question we rebuild the answer from the released telemetry,
rules, and documents on the stated device and window, then accept the
item only if that replay matches the private gold
(Appendix~\ref{app:gold-verification}).
Scored items therefore follow the main-text Stage~3 meaning for the
referenced rule and window.

\subsubsection{Illustrative injector patterns}
\label{app:injector-examples}

\textbf{Consecutive over-current.}
$I_r$ paints three hours of phase current above $I_{\mathrm{rated}}$;
$\operatorname{Eval}$ checks $\mathrm{consec}_{3\mathrm{h}}(i>I_{\mathrm{rated}})$.
A near-miss often lands just below the threshold after $\alpha$-blend.

\textbf{Discrete status fault.}
$I_r$ sets $\textit{operating status}=\texttt{fault}$ at $t_e$;
near-miss status edits follow the discrete $\alpha$ gate.

\textbf{Multi-day slope.}
$I_r$ applies a week-long linear ramp on oil temperature so that
$\mathrm{slope}_{7\mathrm{d}}$ exceeds its Stage 2 parameter;
mutex occupancy spans the full lookback so later events cannot
silently erase the ramp without repair.

\subsection{Grounded Document Generation}
\label{app:document-generation}

This appendix details Stage~4: how the document bundle
$\mathcal{D}_d=(A_d,C_d,F_d)$
is produced from metadata $\mathcal{M}_d$, executable rules
$\mathcal{R}_d$, and temporal evidence $\mathcal{T}_d$
(Eqs.~\eqref{eq:document-definition}--\eqref{eq:document-grounding}).
In the released corpus there are $761$ archives,
$9{,}132$ inspection reports, and $15{,}046$ fault reports.

\subsubsection{Generation principle}
\label{app:doc-principle}

Documents are assembled from facts, not free-form generation.
A program gathers device catalogs, Stage~2 rules, hourly telemetry, and
verified event schedules, then renders fixed tables and structured
sections.
An optional LLM pass fills only marked narrative slots
(briefings, qualitative notes, disposition prose) and may not invent
faults, timestamps, numeric readings, or rule IDs missing from the facts.
Every quantitative claim in $\mathcal{A}_d$, $\mathcal{C}_d$, and $\mathcal{F}_d$ can be recovered from
$(\mathcal{M}_d,\mathcal{R}_d,\mathcal{T}_d)$.

\subsubsection{Device archive $A_d=G_{\mathrm{archive}}(\cdot)$}
\label{app:archive}

One archive is emitted per device.
Program-grounded sections include:
(i)~identity and configuration from $a_d$;
(ii)~a metric dictionary from $\mathcal{K}_d/\mathcal{H}_d$
(name, DB field, unit, normal range / design reference);
(iii)~long-horizon telemetry aggregates from $X_d$
(mean/min/max/last);
(iv)~the applicable rule list from $\mathcal{R}_d$;
(v)~a historical anomaly summary restricted to retained
fault events in $E_d$ that were not skipped.
Manufacture/installation narrative slots may be LLM-filled
under the no-invention constraint.
This yields a persistent equipment dossier aligned with the
same provenance used by later QA.

\subsubsection{Inspection reports $C_d=G_{\mathrm{inspection}}(\cdot)$}
\label{app:inspection}

Inspection calendars come from $L_d$: by default
$12$ bi-monthly anchors at noon, each spawning a
$24$\,h report window $[t,t{+}24\mathrm{h}]$ snapped to the
hourly index---hence exactly $12$ check reports per device
($9132/761$).
Each report includes device/dimension headers, a checklist of
monitored metrics with criteria text drawn from rule
descriptions, an end-of-window snapshot, and the full $24$\,h
metric table from $X_d$.
Point-only rules are re-evaluated inside the window for a
trigger summary.
Near-miss events with $w_{\mathrm{end}}$ no later than the
inspection end are listed explicitly, and the conclusion must
warn when any remain---so inspections can surface
``approaching-threshold'' evidence without labeling a fault.
Checklist wording and in-window point checks draw on the same rule set
$\mathcal{R}_d$ that Stage~2 exports for the device.

\subsubsection{Fault reports $F_d=G_{\mathrm{fault}}(\cdot)$}
\label{app:fault-docs}

Fault reporting is the strongest grounding path.
Episodes are taken from $L_d$: preferably one report per
episode day, attaching all retained fault scenarios whose
$w_{\mathrm{end}}$ falls on that day (primary event $=$ shortest
window); otherwise one report per scenario.
For each matched event $e=(r,w)$, the fact payload records:

\begin{itemize}
  \item rule identity and natural-language judgment text from
  $\mathcal{R}_d$;
  \item named thresholds (parameter table) substituted into
  alarm wording;
  \item input / changed metric fields;
  \item the snapped window $[w_{\mathrm{start}},w_{\mathrm{end}}]$
  and hourly telemetry series for those fields from $X_d$;
  \item measured values at the trigger timestamp.
\end{itemize}

Program sections render alarm tables and time-series panels from these
facts.
LLM narrative (site conditions, impact, field checks, disposition,
root-cause prose) must not contradict the numeric fact tables or invent
new fault modes; causal write-ups cite metric phenomena rather than raw
rule IDs.
Skipped or verification-failed scenarios are excluded.

\subsubsection{Traceability and audit}
\label{app:doc-audit}

Shared provenance makes claims checkable:
archive histories must match retained $E_d$;
inspection windows must land on scheduled inspection days;
fault reports must match device name, rule IDs, and windows in the
event schedule.
A post-hoc auditor verifies these alignments per equipment family.

\subsubsection{Corpus scale}
\label{app:doc-scale}

Per device this is typically $1$ archive, $12$ inspections,
and on the order of $20$ fault reports
(mean $\approx 19.8$), mirroring the Stage 3 event density.

\section{PowerBench Task Construction and Design}
\label{app:taskQuestion}

This appendix details how the agent benchmark is built on the frozen
heterogeneous corpus.
Questions provide only natural-language clues; gold answers are
computed offline from metadata, rules, telemetry, and documents, and
are never exposed through the agent's tools.
The released main set contains $300$ questions:
$60$ TRR , $90$ CRR , and $150$ CoRR.

\subsection{Shared construction principles}
\label{app:task-principles}

\textbf{Clue-only prompts.}
Each question identifies devices by business locators
(e.g., host project, equipment class, configuration fields)
and calendar/report cues, rather than by internal paths or UIDs.
Hard negatives ensure that partial locator matches are insufficient.

\textbf{Executable gold.}
Candidate selection and numerical answers are computed programmatically.
For every accepted item we store a private gold record with the answer
object, evidence atoms, and a step-wise gold trace (device resolve,
reads, metric or rule steps, formatting).
Public files expose only the natural-language question and the required
JSON answer schema.
Appendix~\ref{app:gold-verification} describes the replay checks that
must pass before an item enters the frozen set.

\textbf{Wording pipeline.}
A program first emits an answer-free fact card and a deterministic
business-goal draft; a separate generation model may naturalize the
surface wording.
Programmatic checks reject leakage of hidden execution details,
gold values, or unprotected conditions before a question is frozen.
The evaluated agent never participates in wording generation.

\textbf{Agent interface (evaluation protocol).}
Agents receive five read-only tools over the original corpus:
\texttt{list\_corpus}, \texttt{search\_corpus}, \texttt{read\_file},
\texttt{read\_timeseries}, and a sandboxed \texttt{python} calculator
(no filesystem/network).
Default budgets are $50$ serial tool calls and $300$\,s per question,
with caps on lines/rows/bytes returned per read.
When a budget is exhausted, tools are disabled and the model may
emit one final JSON answer from already collected evidence.

\subsection{Gold verification and frozen-set acceptance}
\label{app:gold-verification}

Corpus-level Stage~3 checks (Appendix~\ref{app:event-stats}) are not the
question oracle.
For each candidate we recompute the answer from the released telemetry,
rules, and documents on the stated device and window, then compare that
replay with the private gold.

\textbf{What is checked.}
Device resolution from business locators; report and timestamp anchors;
metric or rule identity required by the family; window bounds; and the
answer fields (TRR window statistics, CRR threshold policy and replay,
CoRR per-device peaks and the comparative reduction).
Evidence atoms must point to paths that exist in the evaluation scope.

\textbf{Acceptance.}
An item enters the frozen 300-question set only when the replay matches
the private gold.
Items that disagree, leak execution details in the wording, or fail
schema/leakage checks are dropped or regenerated.

\subsection{TRR: Temporal Retrieval and Reasoning}
\label{app:trr}

\textbf{Goal.}
After an inspection event, turn free-text / tabular inspection findings
into post-inspection telemetry analysis.

\textbf{Workflow required of the agent.}
(1)~Resolve the unique device from metadata locators and open its
archive as needed;
(2)~retrieve the inspection report starting on the stated date and
read the end-of-inspection metric excerpt;
(3)~using the metric catalog's normal mean/std, select the numeric
raw metric with largest standardized deviation
$|v-\mu|/\sigma$ at inspection end;
(4)~analyze the subsequent $W$ hourly samples
(default $W{=}72$, exclusive of the end timestamp):
report the selected metric, end-of-inspection $z$-score, unit,
window mean/max, and first time of the maximum.

\textbf{Evidence modalities.}
Inspection report (context + anchor time + candidate readings),
device metadata / metric catalog (semantics + baselines),
and telemetry (post-window evolution).
Gold rejects candidates whose report baseline disagrees with the
CSV at the same timestamp, or whose post-window is degenerate.

\textbf{Answer object (released schema).}
\texttt{selected\_metric}, \texttt{anchor\_deviation\_z}, \texttt{unit},
\texttt{window\_mean}, \texttt{window\_max}, \texttt{first\_max\_time}.

\subsection{CRR: Contextual Retrieval and Reasoning}
\label{app:crr}

\textbf{Goal.}
From a fault narrative, recover operating context, adjust rule
parameters under a stated policy, and re-evaluate the rule on
telemetry.

\textbf{Workflow required of the agent.}
(1)~Resolve the device and the fault report by locator + discovery
date / report number;
(2)~identify the associated executable rule and its original
threshold parameters;
(3)~classify load level from the fault narrative under an explicit
lexicon policy
(high / low / normal; low wins on dual matches) and rescale the
threshold by the corresponding factor
(e.g., $0.8$ / $1.2$ / $0.9$);
(4)~replay the rule on the post-fault evaluation window
(boundary does not reset pre-window consecutive state), reporting
whether it triggers, first trigger time, and trigger-hour count,
together with the inferred load category and effective threshold.

\textbf{Evidence modalities.}
Fault report (event context for parameter choice),
rule specification (expression, consecutive/window conditions,
named thresholds),
and telemetry (replay substrate).
This matches the main-text emphasis: narratives guide
\emph{which} parameters to use; rules + series decide the outcome.

\textbf{Answer object.}
\texttt{load\_category}, \texttt{effective\_threshold}, \texttt{unit},
\texttt{triggered}, \texttt{first\_trigger\_time},
\texttt{trigger\_hours}.

\subsection{CoRR: Comparative Retrieval and Reasoning}
\label{app:corr}

\textbf{Goal.}
Compare multiple independent devices, each with its own documents,
metric definitions, and temporal context, then aggregate on a common
scale.

\textbf{Workflow required of the agent.}
For each aliased device $D_i$ in the prompt
(typically $3$--$5$ units):
(1)~resolve that device from its own locator;
(2)~open its inspection report and metric catalog;
(3)~select its most deviant end-of-inspection raw metric by the same
$|v-\mu|/\sigma$ rule as TRR;
(4)~compute, on its own post-window, the peak absolute standardized
deviation from the catalog normal mean and the first peak time.
Finally compare devices: report per-device selections, identify the
highest-risk device, and return the span between the group's highest
and lowest peak deviations.

\textbf{Evidence modalities.}
Multiple disjoint bundles of
(archive / inspection / catalog / telemetry).
Single-file shortcuts are insufficient; each device must stay tied to
its own evidence.

\textbf{Answer object.}
\texttt{per\_device[]}
(\texttt{alias}, \texttt{selected\_metric},
\texttt{peak\_deviation\_z}, \texttt{first\_peak\_time}),
\texttt{highest\_risk\_device}, \texttt{risk\_score\_span}.

\subsection{Released composition}
\label{app:task-scale}

Pilot runs also tried other family variants
(e.g., fixed-metric post-windows or shifted replays); the reported
matrix freezes the three families above so that paper TRR/CRR/CoRR match
the evaluated items one-to-one.

\subsection{Illustrative questions}
\label{app:task-examples}

We give one English rendering per family (paraphrased from the frozen
Chinese stems).
Gold leaves are shown for orientation; agents never receive them.

\textbf{TRR.}
\emph{Stem (abridged).}
For the device whose archive jointly satisfies
host project $=$ ``Fufeng DC project'', equipment tier $=$ ``secondary
(stability control)'', and hardware version $=$ ``V2.0'', open the
inspection report starting on $2024$-$03$-$01$.
Among numeric raw metrics in the end-of-inspection excerpt that also have
catalog normal mean/std, select the metric maximizing
$|v-\mu|/\sigma$.
Then, over the next $72$ hourly samples after (excluding) inspection end,
report the metric name, end-of-inspection $z$-score, unit, window mean/max,
and first time of the maximum (round to $4$ decimals).

\emph{Gold.}
\texttt{selected\_metric}=\textit{memory utilization};
$z{=}0.9276$; unit $\%$ ;
mean $39.3423$; max $49.1706$; first max at
\texttt{2024-03-03 05:00:00}.

\textbf{CRR.}
\emph{Stem (abridged).}
For the device with host project $=$ ``Xiangjiaba converter station'',
class $=$ ``secondary (AC/DC power)'', role $=$ ``backup'', consider the
fault discovered on $2025$-$08$-$20$ with report id
\texttt{DY-FAULT-20250820-003}.
Rescale the associated rule threshold by load class inferred from the
narrative (high $\times 0.8$, low $\times 1.2$, otherwise $\times 0.9$;
low wins on dual matches) and replay from fault-window end for $72$\,h
without resetting prior consecutive state.
Return load class, effective threshold and unit, whether triggered,
first trigger time, and trigger-hour count.

\emph{Gold.}
\texttt{load\_category}=\texttt{low};
\texttt{effective\_threshold}=$102.0\,^\circ\mathrm{C}$;
\texttt{triggered}=\texttt{false};
\texttt{first\_trigger\_time}=\texttt{null};
\texttt{trigger\_hours}=$0$.

\textbf{CoRR.}
\emph{Stem (abridged).}
Compare three devices $D_1$--$D_3$ given by independent locators and
inspection start dates.
For each device, select its most deviant end-of-inspection raw metric by
$|v-\mu|/\sigma$, then compute the peak absolute $z$-deviation from the
catalog normal mean over the subsequent $72$\,h window and the first peak
time.
Report per-device selections, the highest-risk device, and the span
between the group's highest and lowest peak deviations.

\emph{Gold (excerpt).}
$D_1$ peak $z{=}6.9182$ (output voltage);
$D_2$ peak $z{=}1.2684$;
$D_3$ peak $z{=}1.1652$;
\texttt{highest\_risk\_device}=$D_1$;
\texttt{risk\_score\_span}$\approx 5.753$.

\section{Evaluation Pipeline and Protocol Details}
\label{app:eval-details}

This appendix specifies the agent evaluation loop used for PowerBench: tools, budgets, answer validation, and scoring.
Unless noted, settings match the released main configuration.

\subsection{Interaction loop}
\label{app:eval-loop}

Each question is an independent episode.
The model receives system instructions that require corpus-grounded
reasoning and a user prompt containing the natural-language question
plus the authoritative JSON Schema for the final answer.
Tool calls are executed \emph{sequentially}
(\texttt{parallel\_tool\_calls}=\texttt{false}).
After each call, the tool result is appended to that model's own
transcript; models do not share interaction histories across questions
or across competitors.

\textbf{Read-only tools.}
Five tools are available inside the question's device scope as shown in Table~\ref{tab:agent-tools}

Private gold answers, gold traces, and evidence-atom oracles are never
exposed through these tools.
All access is confined to the scope assigned to the current question
(full-corpus main runs use $N{=}761$ devices).

\begin{table}[t]
\centering
\small
\caption{Default resource limits per question.}
\label{tab:eval-limits}
\begin{tabular}{ll}
\toprule
Limit & Value \\
\midrule
Serial tool-call budget & $50$ \\
Retrieval/reasoning time budget & $300$\,s \\
\texttt{read\_file} max lines / call & $240$ \\
\texttt{read\_timeseries} max rows / call & $500$ \\
Total bytes read / question & $1.5$\,MB \\
\texttt{python} timeout / max stdout & $8$\,s / $12{,}000$ chars \\
Format-correction retries & $1$ \\
\bottomrule
\end{tabular}
\end{table}

\subsection{Budgets and final-answer phase}
\label{app:eval-budget}

The $300$\,s clock covers the tool-assisted retrieval/reasoning phase.
When either the tool-call limit or the time limit is reached, all tools
are disabled and the harness injects a finalization message:
the model must immediately return its best schema-conforming JSON
answer using only evidence already collected.
This finalization turn admits \emph{no} further tool calls, but its
API latency is included in the measured wall-clock time of the run;
hence observed latency may slightly exceed $300$\,s even though
retrieval was cut off at the budget.
Budget exhaustion is recorded independently of answer correctness
(\texttt{tool\_budget\_reached} / \texttt{time\_budget\_reached});
a question can still be scored correct if the final JSON matches gold.

\subsection{Output validation}
\label{app:eval-format}

Model outputs are parsed as JSON and validated against the
question-specific schema (\texttt{jsonschema}).
Markdown fences or non-object payloads count as format failures.
During the normal (tools-enabled) phase, the harness allows at most
one automatic format-correction turn that resends the validation error
and required schema.
After a budget-finalization turn has started, no additional format
retry is granted: a schema-invalid finalization yields
\texttt{format\_error} (or is labeled under time-budget exhaustion if
that path triggered finalization).
Partial structured objects may still be retained for field-level
scoring even when required keys are missing.

\subsection{Scoring}
\label{app:eval-scoring}

\textbf{Joint and field accuracy.}
Let $\hat{y}$ be the predicted answer object and $y^\star$ the
executable gold answer.
Field Accuracy is the fraction of answer leaves that match under
gold-specified absolute numeric tolerances
(timestamps and categorical fields require exact match).
Joint Accuracy is $1$ iff the output is schema-valid \emph{and every}
leaf matches; otherwise $0$.
Macro Joint Accuracy averages Joint Accuracy within each task level
(TRR/CRR/CoRR) and then averages the three levels with equal weight;
Micro Joint Accuracy averages over all $300$ questions.

\textbf{Evidence diagnostics.}
Each gold record defines evidence atoms with acceptable corpus
sources.
From the tool trace we compute retrieval/read/citation recalls over
those atoms (Available Evidence Recall and related variants).
These metrics diagnose whether failures stem from missing evidence
versus incorrect computation; they are not substitutes for Joint
Accuracy.

\textbf{Failure taxonomy.}
Terminal statuses include completed, format error, tool-budget
exhausted, time-budget exhausted, incomplete response, and refusal.
Incorrect completed runs are further typed (retrieval, window
calculation, rule execution, cross-device reduction, format) from
observable score fields for analysis.

\subsection{End-to-end pipeline summary}
\label{app:eval-pipeline}

\begin{enumerate}
  \item \textbf{Build} (offline): select candidates, execute gold,
  emit public questions + private tasks, construct scopes and the
  corpus search index; no evaluated model is involved.
  \item \textbf{Run}: for each (model, question), execute the
  sequential tool loop under Table~\ref{tab:eval-limits}; on budget
  hit, force a no-tool final answer; optionally apply one format retry.
  \item \textbf{Score}: validate schema, compare leaves to gold under
  tolerances, compute Joint/Field Accuracy and evidence recalls,
  aggregate Macro/Micro metrics and budget-exhaustion rates.
\end{enumerate}

This protocol makes PowerBench an agentic stress test of planning under
fixed tool and time budgets, not only of unconstrained chain-of-thought
reasoning.

\section{Metric Definitions}
\label{app:metrics}

This appendix gives the exact definitions used when scoring PowerBench
agent runs. All quantities are computed from the public prediction,
the private gold record, and the recorded tool trace for each question.

\subsection{Answer correctness: JA and FA}
\label{app:metrics-ja}

For question $i$, let $F_i$ be the set of gold answer \emph{leaves}
(scalars after expanding nested objects/arrays, including per-device
fields for CoRR), let $c_{if}\in\{0,1\}$ indicate whether leaf $f$
matches gold, and let $v_i\in\{0,1\}$ indicate that the final output
passes the question-specific JSON Schema.
Numeric leaves use absolute tolerance
$\lvert \hat{y}-y^\star\rvert \le \tau_f$ with
$\mathrm{rel\_tol}=0$
(typical $\tau_f{=}10^{-4}$ for window statistics / $z$-scores;
$\tau_f{=}10^{-6}$ for CRR effective thresholds).
Timestamps and categorical fields require exact equality.
Missing leaves score $c_{if}=0$; the FA denominator remains the
full gold leaf set.

\begin{align}
\mathrm{JA}_i
&=
v_i \prod_{f\in F_i} c_{if},
\\
\mathrm{FA}_i
&=
\frac{1}{|F_i|}\sum_{f\in F_i} c_{if}.
\end{align}

Macro Joint Accuracy weights task families equally:
\[
\mathrm{Macro\text{-}JA}
=
\frac{1}{3}
\sum_{t\in\{\mathrm{TRR},\mathrm{CRR},\mathrm{CoRR}\}}
\frac{1}{|\mathcal{Q}_t|}
\sum_{i\in\mathcal{Q}_t}
\mathrm{JA}_i.
\]
We likewise report mean FA overall and within each
$\mathcal{Q}_t$.
As a secondary aggregate we report Micro-JA
$\frac{1}{N}\sum_{i=1}^{N}\mathrm{JA}_i$ over all $N{=}300$ questions
(unequal family sizes: $60$/$90$/$150$).
Joint Accuracy requires schema validity \emph{and} every leaf correct;
Field Accuracy credits partial progress even when $v_i=0$ or some
leaves fail.

\subsection{Evidence atoms and recall: RER and AER}
\label{app:metrics-evidence}

Each gold record defines a set $\mathcal{A}_i$ of \emph{evidence atoms}.
Atom $a$ specifies a fact the solution needs and a non-empty set of
acceptable corpus sources (raw-relative path + selector, e.g., a
Markdown section/table row, a metric-catalog record, or a CSV field
interval).
Modalities include metadata, documents, and time series.

From the tool trace we mark, for each atom $a$:

\begin{itemize}
  \item $\mathrm{Retrieved}_{i,a}=1$ if any acceptable path for $a$
  appears among paths returned by retrieval-oriented tool results
  (corpus search/listing hits that surface that path);
  \item $\mathrm{Available}_{i,a}=1$ if the atom's content becomes
  usable to the model via a \emph{successful} \texttt{read\_file} /
  \texttt{read\_timeseries} matching an acceptable source, an eligible
  snippet observation, or (in special conditions) pre-provided
  evidence.
\end{itemize}

Per-question recalls are
\[
\mathrm{RER}_i
=
\frac{1}{|\mathcal{A}_i|}
\sum_{a\in\mathcal{A}_i}
\mathbb{I}[\mathrm{Retrieved}_{i,a}],
\qquad
\mathrm{AER}_i
=
\frac{1}{|\mathcal{A}_i|}
\sum_{a\in\mathcal{A}_i}
\mathbb{I}[\mathrm{Available}_{i,a}],
\]
and the reported metrics average over questions:
$\mathrm{RER}=\tfrac{1}{N}\sum_i \mathrm{RER}_i$,
$\mathrm{AER}=\tfrac{1}{N}\sum_i \mathrm{AER}_i$
(likewise within each task family when broken down).

\textbf{Interpretation.}
RER measures whether required sources were \emph{discovered};
AER measures whether required evidence was \emph{fetched into}
the interaction context.
They are complementary, not a strict funnel: a direct read can raise
AER without a prior retrieval hit, so $\mathrm{AER}\not\le\mathrm{RER}$
in general.
High AER with low JA typically indicates reasoning/aggregation failure
after evidence access.

\subsection{Time-budget rate: TBR}
\label{app:metrics-tbr}

Let $\mathrm{TimeHit}_i=1$ if the run marks
\texttt{time\_budget\_reached} (the $300$\,s retrieval/reasoning
budget triggered no-tool finalization).
Then
\[
\mathrm{TBR}
=
\frac{1}{N}\sum_{i=1}^{N}\mathbb{I}[\mathrm{TimeHit}_i].
\]
A time-limit hit does not by itself imply $\mathrm{JA}_i=0$: the
finalization answer is still scored.
(We optionally report an analogous tool-budget reached rate; it is
not part of the primary auxiliary trio in the main text.)

\subsection{Trace metrics: TC, TER, and VS}
\label{app:metrics-trace}

Let $C_i$ be the number of model-issued \texttt{tool\_call} events for
question $i$ (including failed calls), and $E_i$ the number of those
events whose trace records a non-empty error.
Let $\mathrm{SchemaValid}_i\equiv v_i$.
We report
\begin{align}
\mathrm{TC}
&=
\frac{1}{N}\sum_{i=1}^{N} C_i,
\\
\mathrm{TER}
&=
\frac{\sum_{i=1}^{N} E_i}{\sum_{i=1}^{N} C_i}
\quad
\bigl(\text{undefined if }\textstyle\sum C_i=0\bigr),
\\
\mathrm{VS}
&=
\frac{1}{N}\sum_{i=1}^{N} v_i.
\end{align}
Thus TER is a \emph{pooled} error rate over all tool calls in the
evaluation slice, not the mean of per-question error ratios.
VS measures parseability for downstream consumption; a schema-valid
but factually wrong answer still contributes to VS and yields
$\mathrm{JA}_i=0$.

\section{Multi-Device Scaling on CoRR}
\label{app:corr-scaling}

This appendix studies how CoRR performance and resource
use change with the number of target devices under the main evaluation
protocol ($N{=}761$ visible devices; $50$ tool calls / $300$\,s budgets).
The $150$ CoRR questions are partitioned into three equal groups of
$50$ questions with $3$, $4$, and $5$ target devices, respectively.
Because the groups contain \emph{different} questions, the trends below
are descriptive under the evaluation protocol and do not isolate a
causal effect of device count alone.

For GPT-5.6-Sol, Joint Accuracy decreases from $60.0\%$ with three
devices to $40.0\%$ with five, while Field Accuracy decreases from
$83.3\%$ to $66.6\%$
(Table~\ref{tab:corr-scaling}).
The decline therefore extends beyond complete-task success to the
fraction of correctly answered fields: models lose partial credit as
more independent device analyses must be completed and compared.

Partial correctness remains informative when JA is near zero.
Across the same groups, DeepSeek-V4-Pro's FA falls from $20.9\%$ to
$9.1\%$, and Gemini-3.6-Flash's from $18.0\%$ to $3.9\%$, even though
both models' JA is already $\le 4\%$.
Reporting JA and FA together thus characterizes incomplete multi-device
analyses that a single accuracy number would collapse to failure.

\subsection{Resource use and budget pressure}
\label{app:corr-cost}

Resource measurements help interpret these completion patterns.
GPT-5.6-Sol's mean tool calls rise from $26.7$ to $35.6$ as the
device count increases from three to five, while mean input usage
increases from $821.7$K to $1{,}296.6$K tokens and mean latency from
$188$\,s to $286$\,s.
Its time-budget reached rate (TBR) grows from $4.0\%$ to $48.0\%$,
indicating that five-device questions more often exhaust the $300$\,s
retrieval/reasoning budget before a fully tool-supported solution.

Other models show different cost--quality trade-offs.
Qwen3.8-Max uses \emph{fewer} mean tool calls in the five-device group
($29.6$ vs.\ $39.6$ at three devices), but its TBR rises from $64.0\%$
to $98.0\%$ and FA declines.
Lower raw tool consumption therefore does not by itself establish
greater efficiency; it must be read jointly with completion quality and
budget exposure.
DeepSeek-V4-Pro remains near the call ceiling ($\approx 38$--$40$ calls)
with TBR $\ge 86\%$ in all three groups and near-zero JA, consistent
with reliable tool use that still fails at cross-device reduction.

\begin{table}[t]
\centering
\small
\setlength{\tabcolsep}{3.5pt}
\caption{CoRR stratified by number of target devices
($N{=}761$; $50$ questions per cell).
JA/FA/TBR in percent; tool calls and latency are per-question means.
Latency includes the no-tool finalization turn.}
\label{tab:corr-scaling}
\begin{tabular}{llrrrrr}
\toprule
Model & \#dev. & JA & FA & Calls & Lat.\ (s) & TBR \\
\midrule
GPT-5.6-Sol & 3 & 60.0 & 83.3 & 26.7 & 188.2 & 4.0 \\
 & 4 & 46.0 & 72.3 & 30.4 & 227.7 & 10.0 \\
 & 5 & 40.0 & 66.6 & 35.6 & 285.8 & 48.0 \\
\midrule
DeepSeek-V4-Pro & 3 & 2.0 & 20.9 & 38.1 & 495.8 & 96.0 \\
 & 4 & 0.0 & 11.4 & 39.8 & 496.2 & 90.0 \\
 & 5 & 0.0 & 9.1 & 40.4 & 477.5 & 86.0 \\
\midrule
Gemini-3.6-Flash & 3 & 4.0 & 18.0 & 42.3 & 309.6 & 60.0 \\
 & 4 & 2.0 & 8.0 & 41.4 & 312.5 & 76.0 \\
 & 5 & 0.0 & 3.9 & 43.0 & 313.3 & 76.0 \\
\midrule
Qwen3.8-Max & 3 & 0.0 & 4.9 & 39.6 & 331.1 & 64.0 \\
 & 4 & 0.0 & 4.6 & 35.4 & 355.9 & 92.0 \\
 & 5 & 0.0 & 3.4 & 29.6 & 363.8 & 98.0 \\
\bottomrule
\end{tabular}
\end{table}

\subsection{Takeaway}
\label{app:corr-takeaway}

Increasing the number of target devices on CoRR jointly stresses
(i)~per-device local analysis (visible in FA even when JA$\approx 0$),
(ii)~cross-device comparison/reduction (visible in JA), and
(iii)~interaction budgets (visible in calls, tokens, latency, and TBR).
This complements the stage-wise failure modes in the main text:
models may retrieve evidence and execute tools reliably yet still
degrade as more independent device bundles must be completed under a
fixed tool/time envelope.
Together, the stratified CoRR analysis provides a common lens for how
models allocate interaction resources and how much of the requested
multi-device analysis they finish.

\section{Limitations and Threats to Validity}
\label{app:limitations}

\textbf{Synthetic data.}
PowerBench uses an LLM-assisted pipeline with programmatic checks and
web-search audits of taxonomies and inventories
(Appendix~\ref{app:device-audit}).
It is meant to be engineering-plausible and internally consistent, not a
release of confidential utility records.
Results speak to retrieval and reasoning over linked heterogeneous
structure, not to accuracy on any specific utility's private data.

\textbf{Task coverage.}
TRR/CRR/CoRR cover three workflows; they omit many power-system tasks
(e.g., topology reasoning or protection coordination).
Macro-JA averages families with different difficulty.
CoRR strata with $3$/$4$/$5$ devices use different question sets, so
trends are descriptive under the protocol
(Appendix~\ref{app:corr-scaling}).

\textbf{Evaluation stack.}
All models share tools, budgets, prompts, and scoring, but run through
vendor APIs (sometimes via compatible gateways).
TER can mix planning mistakes with adapter or serialization issues, so
we read it as a property of the served model--tool stack.
Hitting the time budget does not by itself mark an answer wrong
(Appendix~\ref{app:eval-details}).

\textbf{Agent design.}
The main matrix uses one shared sequential tool loop with fixed budgets.
We do not claim this is optimal among multi-agent, learned-retriever, or
longer-budget designs.

\textbf{Retrieval-scope runs.}
Appendix~\ref{app:retrieval-scope} uses a $60$-question CoRR subset and
two models at $N\in\{200,400,600,761\}$.
The $N{=}761$ point reuses main-batch runs and may differ slightly in
response-model identifiers, so we do not attribute its JA change to
corpus size alone.

\textbf{Stage~3 checks.}
Eqs.~\eqref{eq:background-generation}--\eqref{eq:event-injection} state
what Stage~3 aims for (quiet background; intended rule fires).
Appendices~\ref{app:bg-eval} and~\ref{app:event-injection} describe how
we implement that with quiet bands, a six-point probe-and-resample
screen, per-event checks of the target rule, and light repair.
We do not claim a checked proof over every hour, and
Eq.~\eqref{eq:event-injection} does not forbid other rules from firing.
Question gold is rebuilt per task window
(Appendix~\ref{app:gold-verification}).

\section{Reproducibility Details}
\label{app:reproducibility}

Artifacts referenced by the Reproducibility Statement ship in the
anonymous repository (full data release after review).

\subsection{Benchmark build and evaluation seeds}
\label{app:repro-bench}

Questions are drafted programmatically, optionally polished by a separate
model, then checked locally for leakage and schema issues.
The evaluated model never sees private gold.
Where the provider exposes it, we use medium reasoning effort; answers
must be plain JSON against the per-question schema, with a plain-text
tool transcript (Appendix~\ref{app:eval-details}).
The frozen set uses build seed $20260906$, scope seed $11$, visible corpus
$N{=}761$, and budgets of $50$ tool calls / $300$\,s.

\subsection{Data-generation settings}
\label{app:repro-datagen}

Device expansion, metric/rule construction, telemetry synthesis, and
document rendering are scripted end-to-end.
Default telemetry seed is $7$, with per-device salted RNGs so identical
seeds do not clone calendars across types.
Background resampling allows up to three retries; default calendars use
$12$ inspection and $8$ episode days with $2$--$4$ alerts per episode.
LLM stages for taxonomy, metrics, and rules call chat APIs compatible
with common OpenAI-style endpoints (released defaults: temperature $0.2$
for instance expansion; temperature $0.0$ for maintenance-metric stages).
Numeric telemetry does not call an LLM.
Document text is template-first; an optional LLM pass only fills marked
narrative slots under fact constraints.
Web-search audits follow Appendix~\ref{app:device-audit}.
Any polishing model used for question wording is configured only at
polish time and never scores answers.

\subsection{How to reproduce scores}
\label{app:repro-run}

\begin{enumerate}
  \item Install the released evaluation package in the documented environment.
  \item Build or load the frozen 300-question bundle with the shipped config.
  \item Run each model through the shared client (API keys via environment
  variables only).
  \item Score with the released field-wise protocol and aggregate Macro-JA,
  FA, RER, AER, TBR, TC, TER, and VS as in Appendix~\ref{app:metrics}.
\end{enumerate}
Published score-file hashes for Section~4 are listed with the analysis
notes in the repository.

\section{Retrieval-Scope Analysis}
\label{app:retrieval-scope}

Beyond device count within a question (Appendix~\ref{app:corr-scaling}),
we study whether enlarging the \emph{visible} device corpus changes CoRR
outcomes.
This analysis uses \emph{existing} runs only: the same $60$ CoRR questions
evaluated for GPT-5.6-Sol and Gemini-3.6-Flash at nested scopes
$N\in\{200,400,600,761\}$ (scope seed $11$), with tools, prompts, gold,
and nominal budgets unchanged.
Questions were stratified by target-device count and wording style
\emph{without} conditioning on model correctness.
Scopes retain targets and designated hard negatives; larger $N$ add
further devices under the nested construction.
The $N{=}761$ endpoint reuses the corresponding main-batch runs on this
$60$-question subset and is marked separately when response-model
identifiers are not fully identical across batches.

\begin{table}[t]
\centering
\small
\caption{Same $60$ CoRR questions at increasing visible corpus size $N$.
JA/FA/RER/AER/TBR in percent. $\dagger$: reused main-batch endpoint.}
\label{tab:retrieval-scope}
\begin{tabular}{llrrrrr}
\toprule
Model & $N$ & JA & FA & RER & AER & TBR \\
\midrule
GPT-5.6-Sol & 200 & 31.7 & 63.7 & 57.4 & 80.6 & 41.7 \\
 & 400 & 33.3 & 66.7 & 57.3 & 84.7 & 41.7 \\
 & 600 & 36.7 & 65.8 & 57.2 & 81.2 & 36.7 \\
 & 761$^\dagger$ & 50.0 & 74.6 & 57.9 & 83.6 & 21.7 \\
\midrule
Gemini-3.6-Flash & 200 & 0.0 & 7.5 & 51.1 & 42.9 & 90.0 \\
 & 400 & 0.0 & 5.2 & 49.8 & 39.6 & 93.3 \\
 & 600 & 0.0 & 6.6 & 48.8 & 37.9 & 81.7 \\
 & 761$^\dagger$ & 0.0 & 8.6 & 50.6 & 39.5 & 71.7 \\
\bottomrule
\end{tabular}
\end{table}

We do \emph{not} observe a consistent JA collapse as $N$ grows among the
newly evaluated scopes ($200$--$600$).
GPT's AER stays in the low-to-mid $80$s and FA in the mid-$60$s; the
paired JA difference between $N{=}200$ and $N{=}600$ is only
$+5.0$ points with a wide exploratory bootstrap interval that includes
zero.
Gemini remains at $0\%$ JA at all four scopes, while nonzero FA and
incomplete AER still provide diagnostic signal.
A plausible design-level reading is that even the smallest scope already
contains designated strong distractors, so further cardinality need not
increase effective ambiguity proportionally under query-based retrieval.
We treat this as a protocol sensitivity check on corpus size, not a causal
claim, and we do not report oracle-evidence ablations here.

\end{document}